\pdfoutput=1

\documentclass[11pt]{article}

\usepackage[]{acl}

\usepackage{times}
\usepackage{latexsym}

\usepackage[T1]{fontenc}
\usepackage{adjustbox}
\usepackage{graphicx}
\usepackage{multirow}
\usepackage{makecell} 
\usepackage{amsmath}
\usepackage{float}
\usepackage{amsfonts}
\usepackage{pifont}
\usepackage{amssymb}
\newcommand{\xmark}{\ding{55}}%

\usepackage[utf8]{inputenc}

\usepackage{microtype}

\usepackage[inline,ignoremode]{trackchanges}

\addeditor{jianshu}

\newcommand{\model}{Z-LaVI}

\title{\model: Zero-Shot Language Solver Fueled by Visual Imagination}

\author{Yue Yang$^{1}$\thanks{~~Work was done during the internship at Tencent AI Lab.} , Wenlin Yao$^{2}$, Hongming Zhang$^{2}$, Xiaoyang Wang$^{2}$, \\\textbf{Dong Yu}$^{2}$, \textbf{Jianshu Chen}$^{2}$ \\
$^{1}$University of Pennsylvania, $^{2}$Tencent AI Lab\\
{\small \tt {yueyang1@seas.upenn.edu}}\\
{\small {\tt \{wenlinyao,hongmzhang,shawnxywang,dyu,jianshuchen\}@global.tencent.com}}}

\begin{document}
\maketitle
\begin{abstract}
Large-scale pretrained language models have made significant advances in solving downstream language understanding tasks. However, they generally suffer from \emph{reporting bias}, the phenomenon describing the lack of explicit commonsense knowledge in written text, e.g., ``\textit{an orange is orange}''.
To overcome this limitation, we develop a novel approach, Z-LaVI, to endow language models with visual imagination capabilities. Specifically, we leverage two complementary types of ``imaginations'': (i) recalling existing images through retrieval and (ii) synthesizing nonexistent images via text-to-image generation. Jointly exploiting the language inputs and the imagination, a pretrained vision-language model (e.g., CLIP) eventually composes a zero-shot solution to the original language tasks. Notably, fueling language models with imagination can effectively leverage visual knowledge to solve plain language tasks. In consequence, Z-LaVI consistently improves the zero-shot performance of existing language models across a diverse set of language tasks.\protect\footnotemark



\end{abstract}

\section{Introduction}

\begin{figure}[!t]
\centering
    \includegraphics[width=7.7cm]{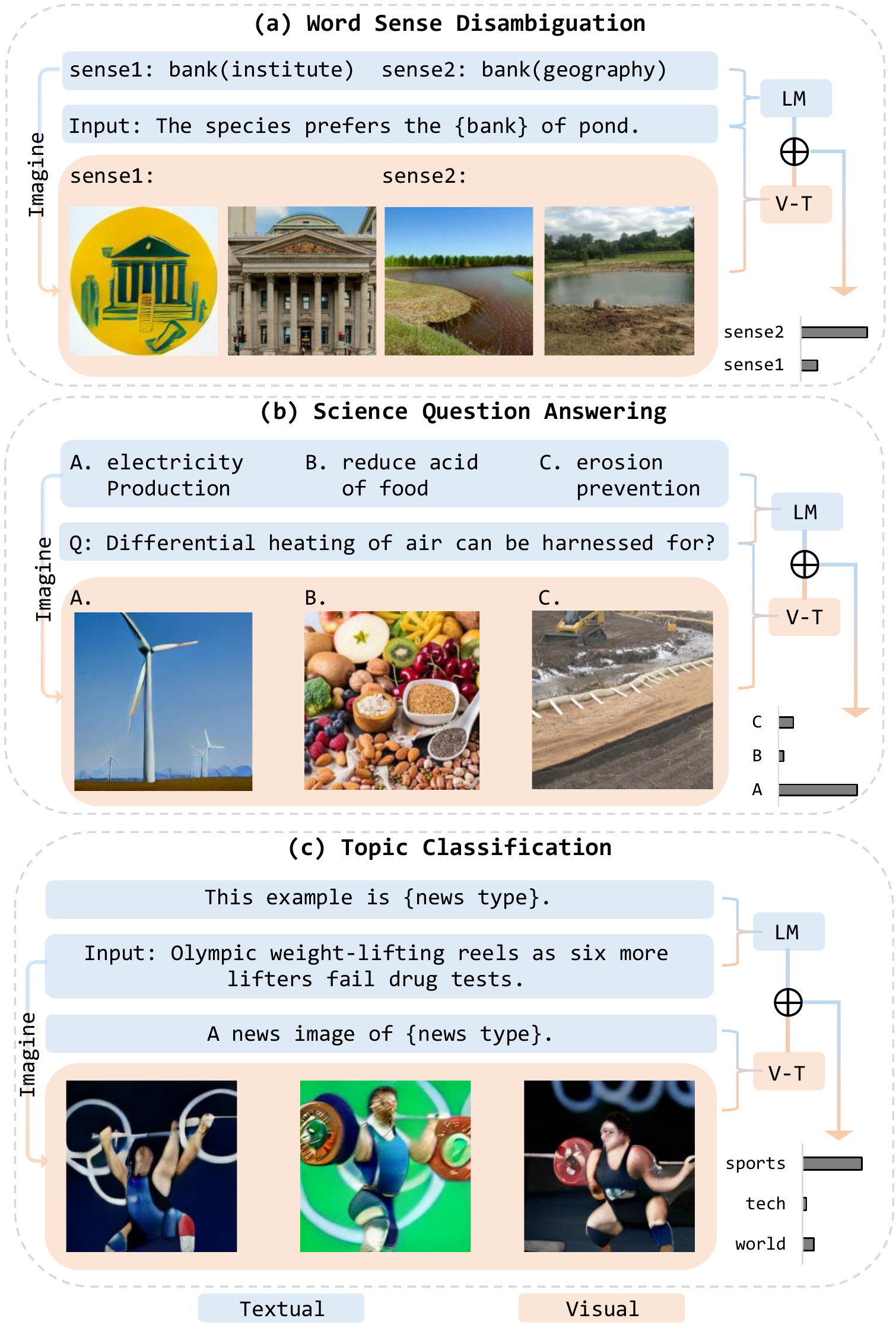}
    \caption{Our system endows language models with two complementary types of visual imagination capabilities: recalling existing images (through retrieval) and synthesizing nonexistent images (via image-to-text generation). They effectively alleviate the reporting bias issue and improves the zero-shot performance for solving \emph{plain} language tasks. We experiment with three types of tasks: (a) Word Sense Disambiguation, (b) Science Question Answering, and (c) Topic Classification.}
    \label{fig:tasks_convertion}
\end{figure}


Large-scale Pretrained Language Models (PLMs) have achieved great success on various Natural Language Understanding (NLU) tasks and even exhibit impressive zero-shot capabilities  without task-specific fine-tunings \cite{radford2019language}. And recent research suggests that such ability improves by further scaling up the model size (e.g., to hundreds of billions of parameters) and the amount of textual pretraining data (to TBs of raw texts) \citep{min2021recent,brown2020language,chowdhery2022palm,kaplan2020scaling}.
However, zero-shot language learners solely trained on texts inevitably suffer from human reporting bias. For example, people tend not to write common or apparent things \citep{grice1975logic}, and the frequency of a certain textual statement does not always correspond to their relative likelihood in the world \citep{gordon2013reporting}.  
Therefore, looking into other modalities to supplement the textual information is crucial.

\footnotetext{Code and data are available at \url{https://github.com/YueYANG1996/Z-LaVI}}

In this paper, we focus on incorporating vision knowledge to facilitate the solution of plain language understanding tasks. Cognitive science has demonstrated that the human vision system is crucial to supplement, interact, and influence the language system \citep{dessalegn2013interaction}. For example, there exists a fast mapping between vision and language in the human language learning process \citep{altmann2004now}.
Inspired by this, we propose a visual imagination framework, \model, to endow any PLMs (e.g., GPT, BERT, BART, etc.) with visual imagination capabilities. 

Specifically, we apply two different types of ``visual imaginations'' to the input texts. Given input text, the first approach \emph{recalls} existing images (e.g., through search engines), and the second one \emph{synthesizes} nonexistent images via text-to-image generation models (e.g., DALL-E \citep{ramesh2021zero}). These two strategies mimic different types of human mental behaviors, i.e., recalling past memories and creative mental image construction. Interestingly, we find that these two mechanisms are highly complementary to each other. 
Our proposed visual imagination module tends to rely more on recalling when input texts are short because their corresponding objects or scenes generally exist and are easy to find. However, when input texts are long and complex, the module is more inclined to create new images.
We develop a unified framework (Figure \ref{fig:tasks_convertion}) that exploits both types of imaginations along with the original textual inputs to compose zero-shot solutions to a broad set of downstream language tasks. Note that our work differs from existing multi-modal tasks such as VQA \citep{antol2015vqa, wu2017visual} or Visual Dialog, \citep{das2017visual}, which have both textual and visual inputs. Instead, we use visual imagination as machinery to facilitate the (zero-shot) solution of pure language tasks.



We show that on a diverse set of language understanding tasks, \model~consistently improves the performance of existing language models of different sizes and architectures. In particular, 
our Z-LaVI with SBERT can achieve a zero-shot F1 score of 87.5\% on the WSD task without fine-tuning, even outperforming BERT-large, which is fine-tuned with three examples per sense, by 2.3\%. Z-LaVI also beats all existing zero-shot models on four Science QA tasks and two Topic Classification tasks by a large margin.
Our analysis demonstrates that Z-LaVI can complement language models and significantly alleviate PLMs zero-shot prediction errors by adaptively executing two visual imagination mechanisms - \textsc{Recall} and \textsc{Synthesis}.


\begin{figure}[!t]
\centering
    \includegraphics[width=7.7cm]{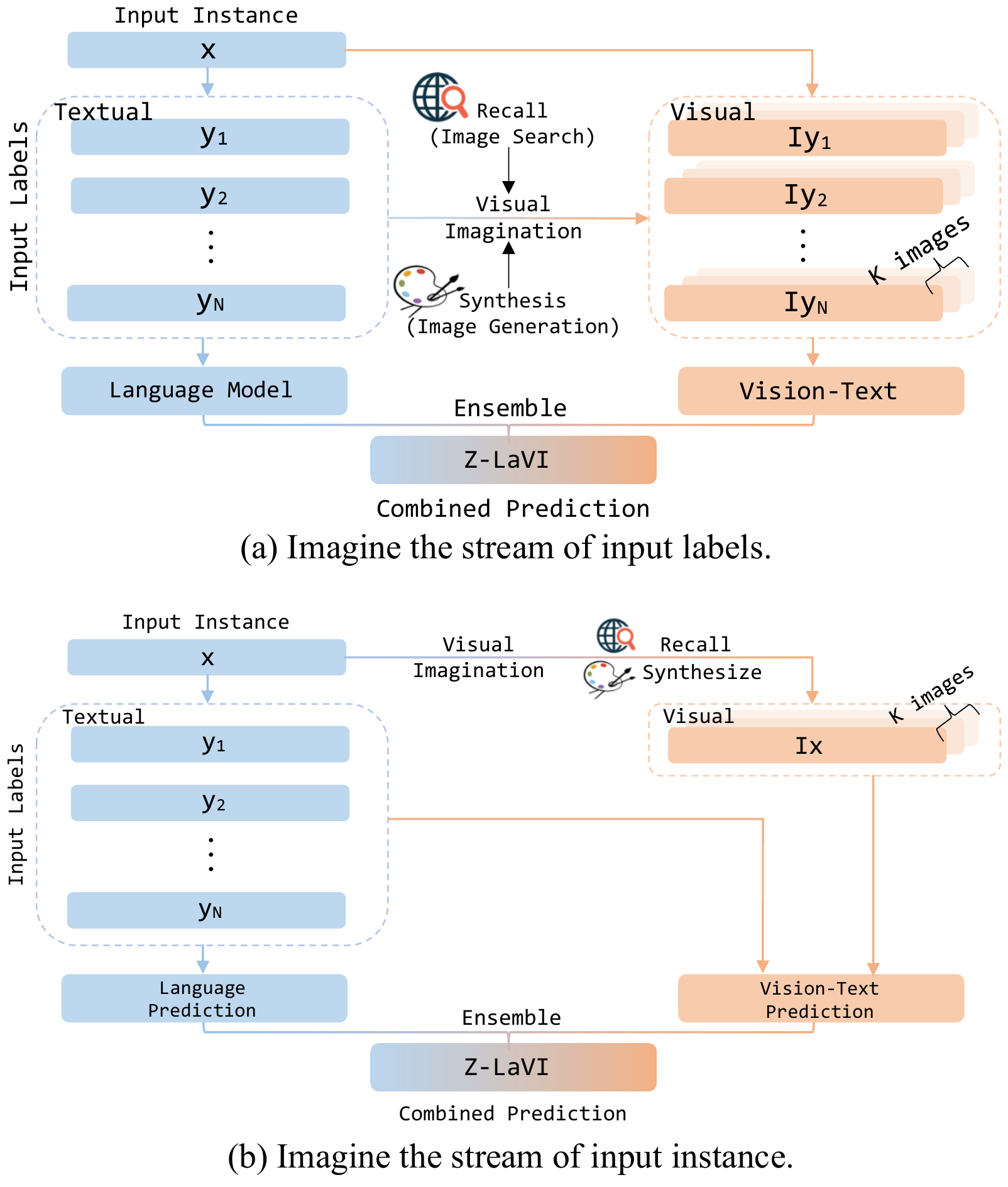}
    \caption{The overview of the proposed Z-LaVI system. Z-LaVI aims to solve the language tasks with two streams of inputs: one stream of label options and another stream of instance to be labeled. Z-LaVI converts one of the streams (either input labels or input instance) into images through visual imagination (\textsc{Recall} and \textsc{Synthesis}) to enable the vision-text model to solve language tasks. We ensemble the language and vision-text models to make the final prediction.}
    \label{fig:model}
\end{figure}

\section{Method}
\subsection{Task Formulation}
To provide a zero-shot solution for language tasks and solve them in a uniform way, we transform different tasks into multi-choice questions, where input stream $x$ and candidate answers stream $y \in \mathcal{Y}$ are provided. The goal is to select the correct answer from $\mathcal{Y}$.
In particular, for word sense disambiguation tasks, $x$ is the instance sentence, and $\mathcal{Y}$ are all possible word senses of the target word; for science question answering tasks, $x$ is the question, and $\mathcal{Y}$ are answer options; for text classification tasks, $x$ is the input sentence, and $\mathcal{Y}$ is the pool of categories.
To make a prediction, the model needs to estimate the plausibility of each tuple $(x, y)$ for all $y \in \mathcal{Y}$ and select the best answer $\hat{y}$.
\begin{equation}
    \hat{y} = \text{argmax}_{y \in \mathcal{Y}} P(y | x).
\end{equation}
\subsection{Language Models for Zero-shot Tasks} \label{sec lm}
We consider three main approaches for employing language models to make zero-shot predictions on language tasks:
\medbreak
\noindent \textbf{Prompt-based Approach} \cite{petroni-etal-2019-language, schick-schutze-2021-exploiting} treats Natural Language Understanding tasks as a cloze test using prompts. For example, we can format question-answering tasks into: 
\medbreak
``\textit{Question}: [$x$]\textit{? The answer is} [$y$].''
\medbreak
We convert the input $(x, y)$ into a sequence of tokens $\textbf{W} = (w_1, ..., w_t, ... w_{t + k}, ..., w_{|\textbf{W}|})$ via a prompt, in which $y = (w_t, ... w_{t + k})$.\footnote{We include the prompts of all tasks in Table \ref{tab: prompt}.} We apply autoregressive language models such as GPT \cite{brown2020language} to calculate the score:
    \begin{align}
    \text{Score}_{\text{La}}(x, y) = \frac{1}{-\frac{1}{k}\sum_{i=0}^{k} \textrm{log}P_{\text{La}} (w_{t+i} | \textbf{W}_{< t + i})}\text{,}
    \nonumber
\end{align}
where $P_{\text{La}}(\cdot)$ denotes the probability given by the language model. Note that we adopt the standard token-length normalization to handle different lengths of answer choices. Finally, we apply softmax to $\text{Score}_{\text{La}}(x, y)$ to obtain the probability of each candidate:
\begin{equation} \label{eq: softmax}
    p_{\text{La}}(y | x) = \frac{e^{\text{Score}_{\text{La}}(x, y)}}{\sum_{y \in \mathcal{Y}} e^{\text{Score}_{\text{La}}(x, y)}}.
\end{equation}

For the prompt-based approach, we select GPT-Neo-1.3B/2.7B \cite{gpt-neo}, GPT-J-6B \cite{gpt-j} and OPT-30B \cite{zhang2022opt} as our models. The GPT-Neo and GPT-J are trained on the Pile dataset \cite{gao2020pile}, which contains 825 GB of English text data. Besides Pile, OPT concatenates the training data of RoBERTa \cite{liu2019roberta} and PushShift Reddit \cite{baumgartner2020pushshift}.
\medbreak
\noindent\textbf{Natural Language Inference (NLI) Approach} 
\cite{yin-etal-2019-benchmarking} propose a textual entailment framework for zero-shot text classification.
The NLI approach considers the input pair $(x, y)$ as a (\textit{premise}, \textit{hypothesis}) pair
to predict the probability that the premise logically entails the hypothesis. 
\begin{equation}
    p_{\text{La}}(y | x) = p(\textsc{Entailment}|(x, y)).
\end{equation}
Note that this approach requires language models to be fine-tuned on (\textit{premise}, \textit{hypothesis}) pairs. 
Here we select RoBERTa-large \cite{liu2019roberta} and BART-large \cite{lewis-etal-2020-bart} fine-tuned on Multi-genre NLI (MNLI) corpus, \cite{williams-etal-2018-broad} consisting of 433k sentence pairs.

\label{sec latent}
\medbreak
\noindent\textbf{Latent Embedding Approach} utilizes an off-the-shelf feature encoder $f_\theta$ to project the input tuple $(x, y$) into a shared latent space and determines their relevance based on 
a distance metric - cosine similarity scores:
\begin{equation}
    \text{Score}_{\text{La}}(x, y) = \text{cos} (f_\theta (x), f_\theta(y)).
\end{equation}
Relevance scores are normalized with softmax (equation \ref{eq: softmax}) to get the final probabilities.

We choose two state-of-the-art sentence encoders, i.e., Sentence-BERT (SBERT) \cite{reimers-gurevych-2019-sentence} and SimCSE, \cite{gao-etal-2021-simcse} as our latent embedding models. For SBERT, we pick the \texttt{all-mpnet-base-v2} checkpoint,\footnote{It is trained on 1B sentence pairs from diverse tasks, including NLI, QA, image captions, etc. The training data have no overlap with our evaluation data, so we still consider this approach as zero-shot.} which achieves the best performance on 14 sentence embedding datasets.\footnote{\href{https://www.sbert.net/docs/pretrained_models.html}{https://www.sbert.net/docs/pretrained\_models.html}} For SimCSE, we choose the best fully unsupervised model \texttt{unsup-simcse-roberta-large}.

\subsection{Language with Visual Imagination}
\medbreak
\noindent\textbf{Visual Imagination} aims to convert either $x$ or $y$ (depending on the task) in the textual input tuple $(x, y)$ into an image. For WSD and QA tasks, we imagine the candidate options $y$. While for topic classification tasks, we imagine the instance sentence $x$. Here we illustrate our method through the example of imagining $y$. We propose two imagination mechanisms: 1) \textsc{Recall} and 2) \textsc{Synthesis}.
\medbreak
\noindent 1) \textsc{Recall}: We use the text input to query Bing Image Search\footnote{We use the \href{https://learn.microsoft.com/en-us/azure/cognitive-services/bing-image-search/overview}{Azure Bing Image Search} API.} to recall the corresponding images. We set a maximum number of images for each query. When only limited images are available for some queries, we download all of them.
\medbreak
\noindent 2) \textsc{Synthesis}: We adopt DALL$\cdot$E \cite{ramesh2021zero}, a text-to-image generation model pretrained on image-caption pairs, to synthesize images. DALL$\cdot$E constructs a codebook $\mathcal{V}$ using a discrete variational autoencoder (dVAE) \cite{rolfe2016discrete} to map the image into tokens concatenated with the caption's text tokens. DALL$\cdot$E models the joint distribution over the text and image tokens with an autoregressive transformer. During inference, DALL$\cdot$E feeds the text tokens $y$ into the transformer and generates a sequence of image tokens $(v_1, v_2, ..., v_m)$, where an image token $v_i$ is predicted based on the previous ones:
\begin{equation}
    v_i = \text{argmax}_{v \in \mathcal{V}}p(v | y, v_{<i}),
\end{equation}
in which, $\mathcal{V}$ is the visual codebook. After we generate enough image tokens, we decode the tokens into images by looking up the vectors in the dVAE codebook to construct the pixels. 

We iterate the \textsc{Synthesis} process multiple times and combine with the images from \textsc{Recall} to collect a set of $K$ images $\{I_y^k|i=1, ..., K\}$ for each textual input $y$.\footnote{The two methods will produce more than $K$ images, and we select the top-$K$ based on their similarity with the text input calculated by CLIP.}
\medbreak
\noindent\textbf{Vision-Text model for Zero-shot language tasks.} After transferring an input stream into images, we modify a plain language task into a multimodal task. Thus we can apply vision-text models to solve the problems. We choose CLIP \cite{radford2021learning} as our vision-text model, which is pre-trained on 400M image-caption pairs with the contrastive learning strategy. 

CLIP has a text encoder $f_{\mathrm{T}}$ and a visual encoder $f_{\mathrm{V}}$, which can project text and image into the shared latent space. Similar to the latent embedding approach described in \ref{sec latent}, we aggregate the $K$ images collected previously and use CLIP to compute the relevance score of $(x, y)$:
\begin{equation} \label{eq VI score}
    \text{Score}_{\text{VI}}(x, y) = \frac{1}{K} \sum_{k=1}^K \text{cos} (f_{\mathrm{T}} (x), f_{\mathrm{V}}(I_y^{k})),
\end{equation}
and we obtain a probability distribution through softmax (over $y$):
\begin{equation}
    p_{\text{VI}}(y | x) = \text{softmax}\left(\text{Score}_{\text{VI}}(x, y)\right).
\end{equation}
\medbreak
\noindent\textbf{Ensemble Language and Vision Prediction.} Our system is designed for zero-shot tasks without labeled data to learn weights to ensemble the two models. Therefore, we adopt a weighted sum as the late fusion over the final output distributions of the language and multi-modal models:
\begin{equation} \label{eq ensemble}
    p_{\text{LaVI}}(y|x) = (1 - w) \cdot p_{\text{La}}(y|x) + w \cdot p_{\text{VI}}(y|x),
\end{equation}
where we design a heuristic function to calibrate the weight $w$ based on the relative size between the vision-text model and the language model:
\begin{equation} \label{eq: ensemble weight}
    w = \text{sigmoid}\left(\frac{\mathcal{P}_{\text{VI}}}{\mathcal{P}_{\text{La}}}\right),
\end{equation}
where $\mathcal{P}_{\text{VI}}$ and $\mathcal{P}_{\text{La}}$ are the number of parameters of the models. We hypothesize that when the language model's size increases, it will encode more knowledge and thus rely less on the vision model. The number of parameters of each model and their corresponding weight is listed in Table \ref{tab n parameters}.

\section{Experimental Setup}
\subsection{Datasets}
\begin{table}[!t]
\centering
\resizebox{7.7cm}{!}{
\begin{tabular}{cccc}
\Xhline{3\arrayrulewidth}  
\textbf{Task}    & \textbf{Dataset}     & \textbf{Split} & \textbf{\# samples} \\ \hline
WSD & CoarseWSD-20 & test  &  10,196\\ \hline
\multirow{4}{*}{\begin{tabular}[c]{@{}c@{}}Science \\QA\end{tabular}} & QASC        & dev   & 926        \\
                            & SciQ        & dev   & 1,000      \\
                            & ARC-E       & dev   & 570        \\
                            & ARC-C       & dev   & 299        \\ \hline
\multirow{2}{*}{\begin{tabular}[c]{@{}c@{}}Text \\  Classification\end{tabular}} & AG-News     & test  & 7,600      \\
                            & Situation   & test  & 1,789     \\\Xhline{3\arrayrulewidth}  
\end{tabular}
}
\caption{Dataset statistics for the three tasks.}
\label{tab: data stat}
\end{table}
We evaluate our methods on six datasets of three tasks. Table \ref{tab: data stat} shows dataset statistics.
\medbreak
\noindent\textbf{CoarseWSD-20} \cite{loureiro2021analysis} is a coarse-grained WSD dataset built from Wikipedia. The dataset consists of 20 nouns with 2-5 senses per noun (53 senses in total). Each sense is associated with a definition which is the first sentence on its Wikipedia page. CoarseWSD guarantees that every sense has test instances in the test set. On average, each sense has 192 test instances.
\medbreak
\noindent\textbf{QASC} \cite{khot2020qasc} is a multi-hop, 8-way choice question answering dataset collected by decomposing sentences about scientific facts. We report the performance of the development set, which contains 926 questions.
\medbreak
\noindent\textbf{SciQA} \cite{welbl-etal-2017-crowdsourcing} is a dataset of 4-way multiple-choice science exam questions spanning from elementary to college-level covering chemistry, biology, physics, etc. We evaluate the development set with 1,000 questions.
\medbreak
\noindent\textbf{ARC} \cite{Clark2018ThinkYH} consists of 7,787 natural, grade-school level science questions. The ARC dataset is split into easy (ARC-E) and challenge (ARC-C), where questions in the challenge set contain the ones that simple retrieval or word correlation methods cannot answer correctly. We evaluate the development sets of ARC-E and ARC-C, which contain 570 and 299 questions, respectively.
\medbreak
\noindent\textbf{AG News} \cite{zhang2015character} is a news topic classification dataset, and each sentence is associated with one of the four news types: \textit{word}, \textit{sports}, \textit{business}, and \textit{technology}. We run our models on the 7,600 examples in the test set.
\medbreak
\noindent \textbf{Situation} \cite{mayhew2018university} is a event-type classification task. The dataset has 12 events: \textit{need water}, \textit{need infrastructure}, \textit{crime violence}, etc. The original task on this dataset is multi-label classification and has an \textit{out-of-domain} class. As the multi-label prediction requires a fine-tuned threshold to determine the predictions and is thus not suitable for zero-shot models, we remove those examples with more than one label and ones with the \textit{out-of-domain} label, resulting in 1,789 instances.

\subsection{Baselines}
Aside from the zero-shot language models described in the section \ref{sec lm}, we also evaluate on a random baseline and compare with previous work.

For CoarseWSD-20, we compare with the BERT-large few-shot (1-shot/3-shot per sense) results reported in \citet{loureiro2021analysis}. 

For QA tasks, we include the Information-Retrieval (IR) solver~\cite{clark2016combining}, which combines the question and option as a query and sends it to a search engine to check if they are explicitly written in some corpus. We also choose SMLM \cite{banerjee-baral-2020-self} as another baseline - a RoBERTa-large model fine-tuned on triplets extracted from knowledge graphs such as ATOMIC \cite{sap2019atomic}.

We compare topic classification with the TE-wiki \cite{ding-etal-2022-towards-open}, the state-of-the-art model on zero-shot topic classification trained on a dataset collected from Wikipedia. 

\begin{table*}[!t]
\centering
\small
\begin{tabular}{lccccccccc}
\\\Xhline{2\arrayrulewidth}    
\multirow{2}{*}{\textbf{Model}} & \multirow{2}{*}{\textbf{\# Param.}} & \multicolumn{2}{c}{\textbf{QASC}} & \multicolumn{2}{c}{\textbf{SciQ}} & \multicolumn{2}{c}{\textbf{ARC-E}} & \multicolumn{2}{c}{\textbf{ARC-C}} \\ \cline{3-10} 
& & Original  & \model      & Original      & \model      & Original       & \model      & Original      & \model      \\ \hline
Random       & - & 12.5    &  -   &  25.0   &  -  &    25.0  & -  & 25.0    & -\\
IR Solver$^{\dagger}$ & - &  18.6   &  -  &  -    &  -  &  30.4    & -  &  20.3   & - \\
SMLM$^{\dagger}$ & 355 M &  26.6   & - &  -   & - &   33.4   & - &  28.4   & - \\ 
RoBERTa-L-mnli*  & 355 M & 19.3   & \underline{27.2}    &  44.7 & \underline{51.3}  &  48.4  & \underline{51.8} &  34.4  & 33.4\\
BART-L-mnli*   & 400 M & 21.7    & \underline{27.3}    &  48.8   &  \underline{51.0}   & 54.7     & \underline{56.1} &  \textbf{36.5}   & \textbf{36.5}\\
GPT-Neo-1.3B    &  1.3B   & 29.3    & \underline{37.4}    &  57.5   & \underline{60.8} &  46.3 & \underline{49.8} & 27.4 & 26.1\\
GPT-Neo-2.7B    &  2.7B   & 29.6    & \underline{39.6}    &  64.0   & \underline{64.9}    &  49.6   &  \underline{51.9}   &  31.8    & 30.4\\
GPT-J-6B     &  6B  &  36.3   & \underline{42.0}    &  73.2   & \underline{73.7} &  55.1    & \underline{57.2} & 34.8   & 34.1 \\
OPT-30B & 30B & 39.7 & \underline{\textbf{42.1}} & 72.7 & \underline{\textbf{74.0}} & 58.2 & \underline{\textbf{59.5}} & 34.8 & 34.1\\
SimCSE     & 355M  & 30.8    &  \underline{33.2}   &  42.6   & \underline{48.6}  &  43.3    & \underline{49.3}  &  26.4   & 24.7 \\
SBERT*     & 110M  &  36.7   & \underline{38.6}    &  57.7   & \underline{58.5}    & 54.4     & \underline{56.0} &    30.1 & 27.1\\\hline
\model~w/o LM  & 150M  &  -   & 32.7   &  -   &  49.5   &  -    & 50.2 &   -  &  26.7 \\\Xhline{2\arrayrulewidth}    
\end{tabular}
\caption{Zero-shot performance on Science QA tasks. \model~represents the performance with our Visual Imagination. \model~(w/o LM) is the model that only uses vision-text prediction. The best-performed number for each metric is \textbf{bolded}. The numbers are \underline{underlined} if the original performance is improved with \model. The models with * use labeled data for pre-training. The models with $\dagger$ indicate the results are from previous work.}
\label{tab: qa}
\end{table*}

\subsection{Evaluation Metrics}
We report the accuracy of all question-answering and topic-classification datasets. For CoarseWSD-20, we compute each word's accuracy and F1 score and take the mean score of all 20 words.

\subsection{Implementation Details}
\textbf{Image Collection} We adopt Bing Image Search to \textsc{Recall} images. And for image \textsc{Synthesis}, we utilize the newly released DALL$\cdot$E-mini\footnote{\href{https://github.com/borisdayma/dalle-mini}{https://github.com/borisdayma/dalle-mini}, and we use the DALL$\cdot$E-mega checkpoint.} which chooses VQGAN \cite{esser2021taming} as the image encoder/decoder and BART \cite{lewis-etal-2020-bart} as the autoregressive transformer. For every textual input, we obtain 100 images from each of the two methods. The 200 images are sorted using CLIP based on their similarity with the text input. We preserve each text input's top-10 images ($K = 10$) and feed them into the equation \ref{eq VI score} to calculate the vision-text probabilities.
\medbreak
\noindent \textbf{Model Implementation} The GPT-style and NLI-based language models are built on top of the huggingface API.\footnote{\href{https://huggingface.co}{https://huggingface.co}} For NLI models, we use the recently released zero-shot classification pipeline.\footnote{\href{https://huggingface.co/facebook/bart-large-mnli}{https://huggingface.co/facebook/bart-large-mnli}} We use the official release of SBERT\footnote{\href{https://www.sbert.net}{https://www.sbert.net}} and SimCSE\footnote{\href{https://github.com/princeton-nlp/SimCSE}{https://github.com/princeton-nlp/SimCSE}} to implement the latent embedding approach. The CLIP model is adapted from the OpenAI's public repo,\footnote{\href{https://github.com/openai/CLIP}{https://github.com/openai/CLIP}} and we select the ViT/B32 as the image encoder. The experiments were run on 3 $\times$ 8 NVIDIA V100 32GB, which can generate 24 images in 5 seconds. The majority of the running time of our model is image generation. In total, we employ DALL$\cdot$E-mini to generate approximately 1.8M images which take around 104 hours.

\begin{table}[!t]
\centering
\resizebox{7.5cm}{!}{%
\begin{tabular}{lcccc}
\Xhline{3\arrayrulewidth}  
\multirow{2}{*}{\textbf{Model}} & \multicolumn{2}{c}{\textbf{Accuracy}}             & \multicolumn{2}{c}{\textbf{F1}}       \\ \cline{2-5}
   & \multicolumn{1}{l}{Orig.} & \multicolumn{1}{l}{\model} & \multicolumn{1}{l}{Orig.} & \multicolumn{1}{l}{\model} \\ \hline
Random                             &   41.3   &   -   &  36.7    &   -   \\
BERT-L-1shot$^{\dagger}$*          &  77.6    &  -    &  71.2    &   -   \\
BERT-L-3shot$^{\dagger}$*          &  89.3    &  -    &  85.2    &   -   \\ 
RoBERTa-L-mnli*       &  80.4    &  \underline{83.0}    &  74.4    &  \underline{78.1} \\
BART-L-mnli*          &  80.2    &  \underline{82.4}    &  74.8    &  \underline{77.9} \\
GPT-Neo-1.3B          &  84.7    &  \underline{88.4}    &  78.3    & \underline{84.6}  \\
GPT-Neo-2.7B          &  86.7    &  \underline{88.9}    &  81.5    & \underline{85.3}  \\
GPT-J-6B              &  84.1    &  \underline{88.5}    &  79.3    & \underline{84.8}  \\
OPT-30B               &  84.4    &  \underline{88.8}    &  80.4    & \underline{85.1}  \\
SimCSE                &  85.1    &  \underline{89.7}    &  78.9    &  \underline{86.0}    \\
SBERT*                &  87.8    &  \textbf{\underline{90.6}}    &  83.3    &  \textbf{\underline{87.5}}    \\ \hline
\model~w/o LM                  &   -   &   87.7   &   -   &   83.8   \\
\Xhline{3\arrayrulewidth}  
\end{tabular}
}
\caption{Zero-shot performance on CoarseWSD-20.}
\label{tab:wsd}
\end{table}
\vspace{-.2cm}

\section{Evaluation}
\subsection{Main Results}
\paragraph{Z-LaVI boosts the performance of language models.} Table \ref{tab: qa}, \ref{tab:wsd} and \ref{tab: tc} show results on seven datasets of three tasks. Each dataset has two results columns: the original performance of the language models and the ensembled performance by adding our \model~model. We observe that in most cases, \model~consistently improves the performance of different language models. Especially in the WSD task, our Z-LaVI with SBERT can outperform the BERT-large fine-tuned with 3-shots of each sense. \model~also significantly enhances the language models on topic classification task where the best language model with \model~beats the SOTA zero-shot topic classification model TE-wiki by 2.8\%. For science QA tasks, we can see \model~improves on QASC, SciQ, and ARC-E, but it struggles on the ARC-C, and adding \model~degrades the performance of a few language models. This is because the ARC-C questions are designed to be hard to answer using retrieval or correlation, and \model~uses CLIP, which is pre-trained on the image-text correlation only.
Figure \ref{fig:qa_examples}~(b) shows an example that needs multi-hop reasoning where \model~fails to answer correctly.

\paragraph{\model~without language model is a strong baseline.} Surprisingly, we also find that \model~w/o language model performs well on plain language tasks. In some datasets, such as QASC, Coarse-WSD, and topic classification tasks, \model~w/o LM outperforms the language models without fine-tuning on the downstream datasets (e.g., SimCSE, GPT-Neo-1.3B/2.7B). This indicates that the vision-text model pretraining on image-caption pairs learns the knowledge that can be leveraged to solve single modality tasks.

\begin{figure}[!t]
\centering
    \includegraphics[width=7.7cm]{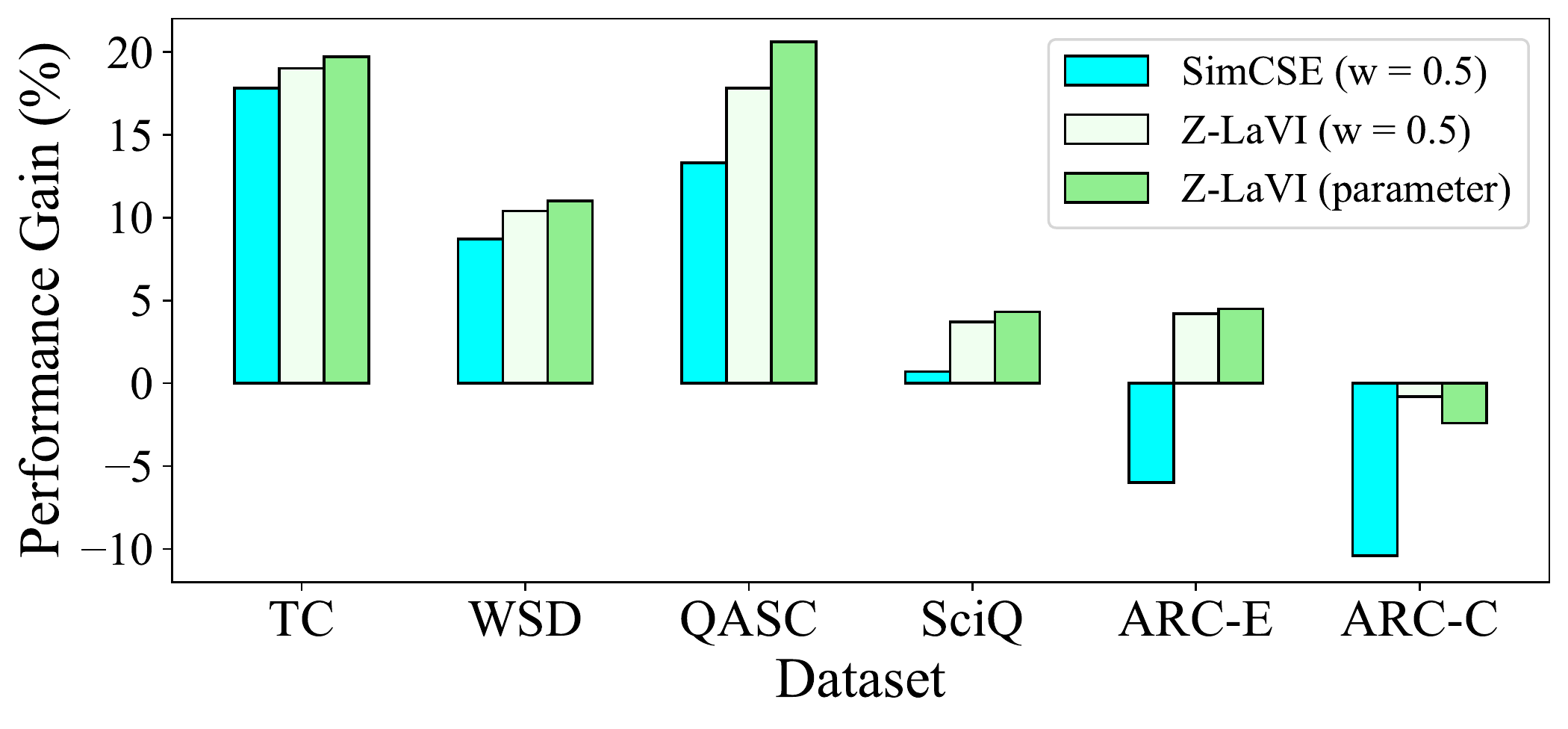}
    \caption{The average performance gain on each dataset. \model~(parameter) stands for accounting models' number of parameters (Equation \ref{eq: ensemble weight}) to adjust the weights.}
    \label{fig:pg}
\end{figure}

\paragraph{Ensembling two language models is not as good as \model.} To verify the effectiveness of using visual knowledge, we replace the visual imagination of \model~with another language model - SimCSE. 
We select SimCSE here because SimCSE is trained fully unsupervised and has the same contrastive learning objective as CLIP. We define the performance gain (\texttt{PG}) of model $\mathcal{M}_{1}$ (i.e., SimCSE) on top of model $\mathcal{M}_{2}$ by computing the relative improvement of the ensemble model $\texttt{Ens} (\mathcal{M}_{1}, \mathcal{M}_{2})$ performance over the original model $\texttt{Orig}(\mathcal{M}_{2})$.
\begin{table}[!t]
\centering
\resizebox{7.7cm}{!}{
\begin{tabular}{lcccc}
\Xhline{3\arrayrulewidth}  
\multirow{2}{*}{\textbf{Model}} & \multicolumn{2}{c}{\textbf{AG-News}} & \multicolumn{2}{c}{\textbf{Situation}} \\ \cline{2-5} 
    & Orig.  & \model    & Orig. & \model  \\ \hline
Random       &   25.0  &   -   &    9.1   &    -   \\
TE-wiki$^{\dagger}$*     &   79.6  &   -   &     -    &    -   \\
RoBERTa-L-mnli*  &   81.2  &  \underline{81.7}     &  40.7  &  \underline{41.8}  \\
BART-L-mnli*  &   81.9  &  \underline{\textbf{82.4}}     &  40.5  &  \underline{41.1}  \\
GPT-Neo-1.3B &   59.1  &  \underline{72.9}     & 17.8     & \underline{38.5}   \\
GPT-Neo-2.7B &   59.1  &  \underline{74.5}     &  13.6   &   \underline{35.2} \\
GPT-J-6B     &   61.0  &   \underline{73.8}    &  21.9   &  \underline{38.8}  \\
SimCSE       &   58.1  &  \underline{73.1}     &  42.1   &  \underline{44.4}  \\
SBERT*       &   77.8  &  \underline{82.2}     & 42.6    &  \underline{\textbf{46.6}}     \\ \hline
\model~w/o LM       &    -     &  71.6     &  -   &  33.4  \\ \Xhline{3\arrayrulewidth}
\end{tabular}
}
\caption{Zero-shot Performance on Topic Classification.}
\label{tab: tc}
\end{table}
\vspace{-.2cm}

\begin{equation}
    \texttt{PG}(\mathcal{M}_{1}, \mathcal{M}_{2}) = \frac{\texttt{Ens} (\mathcal{M}_{1}, \mathcal{M}_{2}) - \texttt{Orig} (\mathcal{M}_{2})}{\texttt{Orig}(\mathcal{M}_{2})}
\end{equation}
We include all the language models (exclude SimCSE) in the set $\mathbb{M}$\footnote{6 models in total, which are GPT-Neo-1.3B/2.7B, GPT-J-6B, RoBERTa-L-mnli, BART-L-mnli, and SBERT.} and calculate the average performance gain on a dataset by:
\begin{equation}
    \texttt{avg-PG}(\mathcal{M}_{1}) = \frac{1}{|\mathbb{M}|} \sum_{\mathcal{M} \in \mathbb{M}} \texttt{PG}(\mathcal{M}_{1}, \mathcal{M})
\end{equation}
For fair comparison, we fix the ensemble weight $w = 0.5$ in equation \eqref{eq ensemble} for both SimCSE and \model. We also include the \model~with dynamic ensemble weight controlled by equation \eqref{eq ensemble}. The performance gain of SimCSE and Z-LaVI on all six datasets is shown in Figure \ref{fig:pg}. We observe that Z-LaVI consistently has higher performance gain than SimCSE across all datasets, demonstrating that the visual information provided by \model~
complements language models more hence boosts more on performance.
Additionally, \model~with dynamic weights perform better than simply setting the weight to 0.5. 

\begin{figure}[!t]
\centering
    \includegraphics[width=6cm]{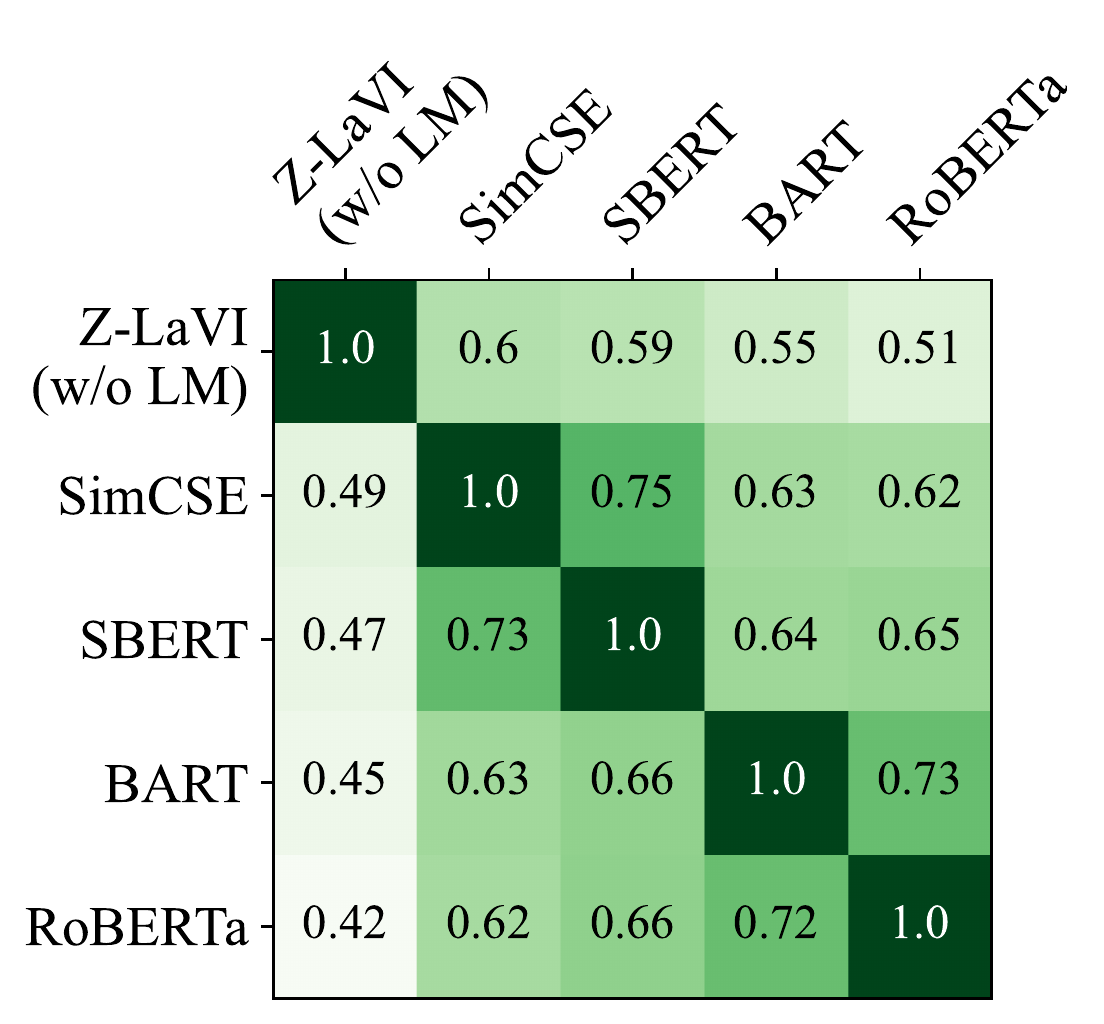}
    \caption{The overlap of correctly predicted examples between each pair of models in the Situation dataset.}
    \label{fig:overlap}
\end{figure}

\subsection{Analysis}
\begin{figure}[!t]
\centering
    \includegraphics[width=7.7cm]{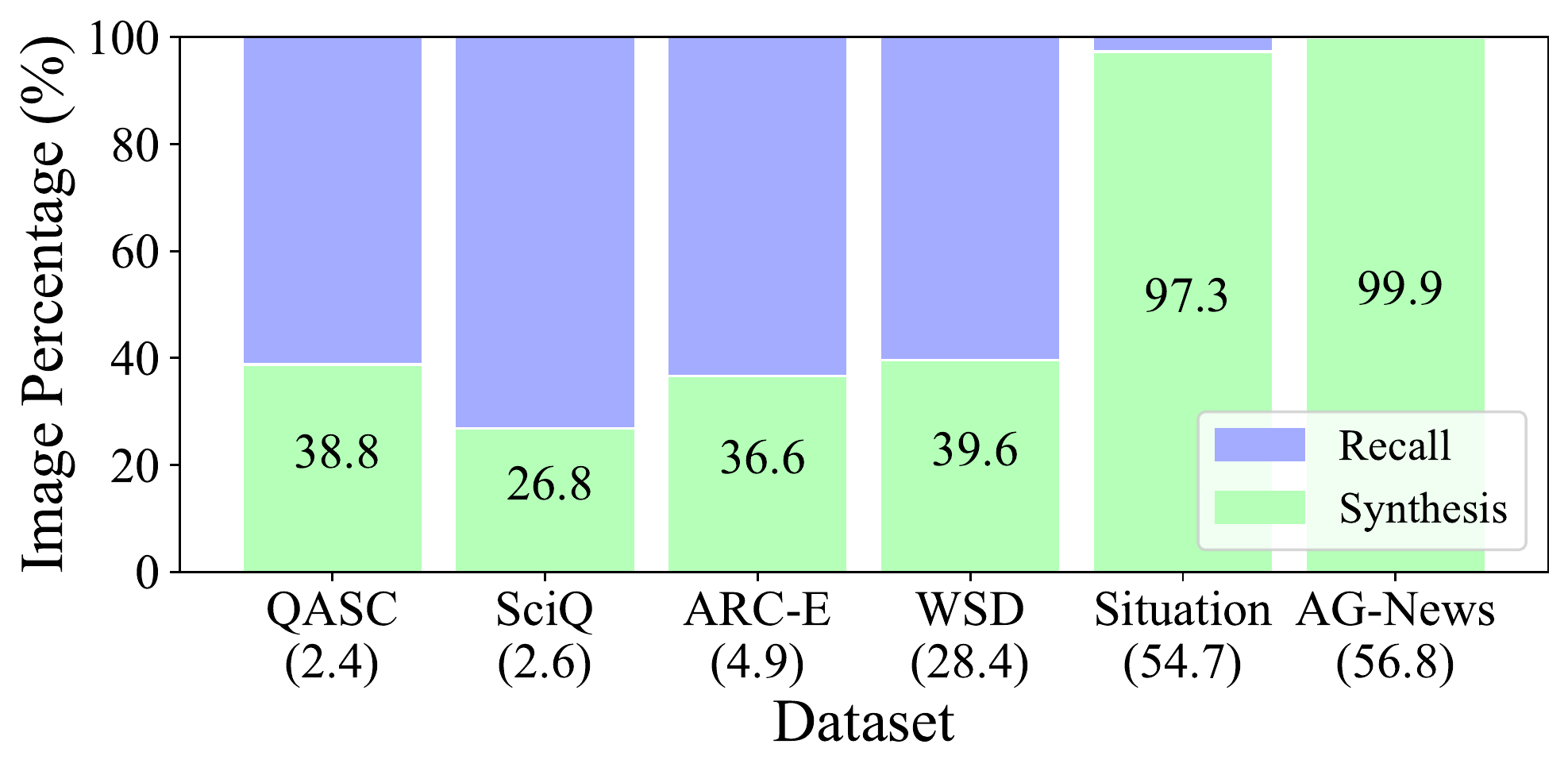}
    \caption{The percentage of recall and synthesis images within the top 10 images of each dataset. The average token length for each dataset is given on the x-axis.}
    \label{fig:image_type}
\end{figure}

\begin{figure}[!t]
\centering
    \includegraphics[width=7.7cm]{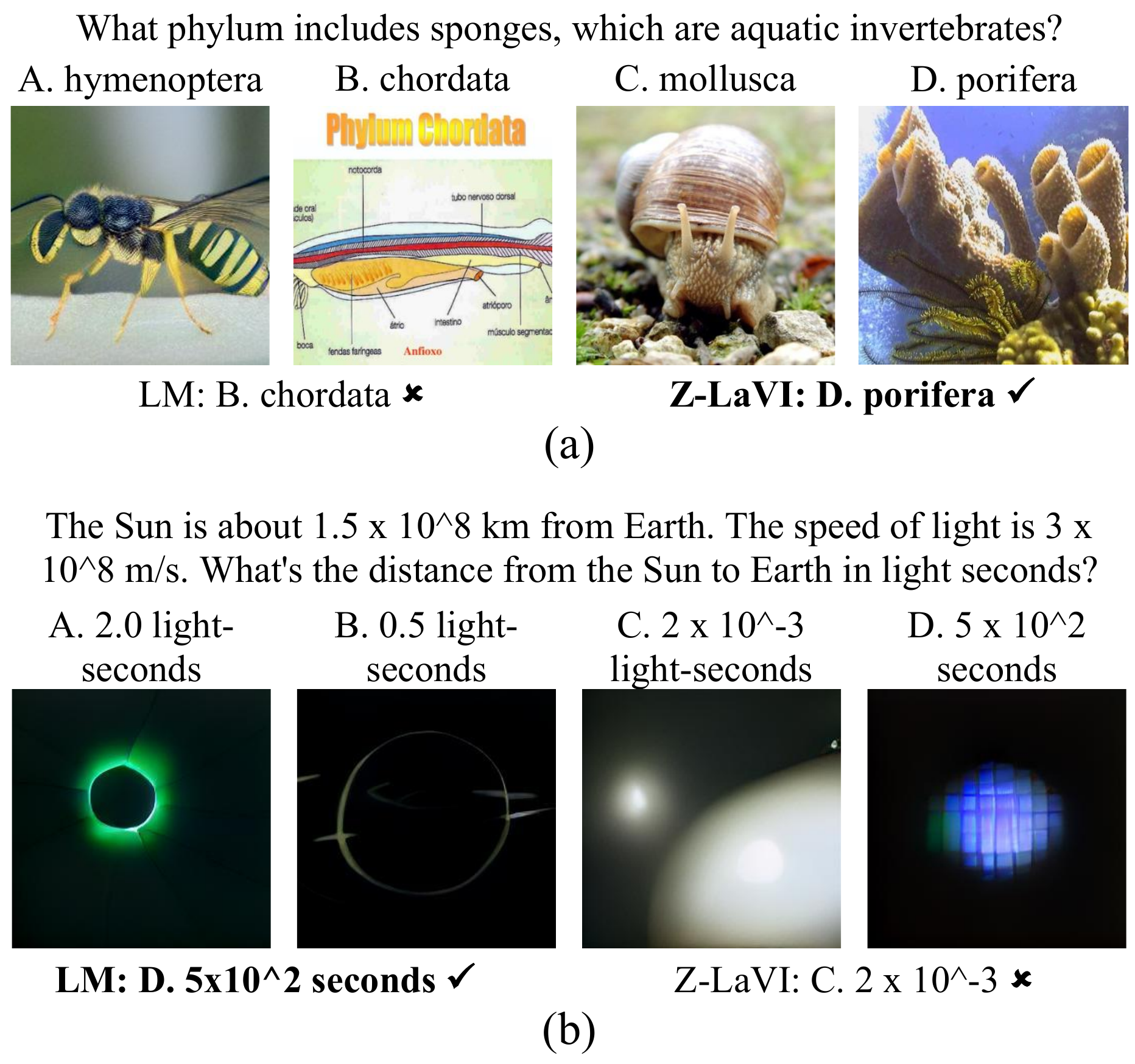}
    \caption{Qualitative examples from (a) SciQ and (b) ARC-C. \model~can successfully answer the questions which can be solved by correlation. \model~ fails to answer the question that requires multi-hop reasoning.}
    \label{fig:qa_examples}
    \vspace{-.15cm}
\end{figure}

\paragraph{Vision and Language models behave differently.} 
We define the overlap of correctly predicted examples between two models as: 
\begin{equation}
    \texttt{overlap}(\mathcal{M}_1, \mathcal{M}_2) = \frac{|S_{\mathcal{M}_1^*} \cap S_{\mathcal{M}_2^*}|}{|S_{\mathcal{M}_1^*}|}
\end{equation}
where $S_{\mathcal{M}^*}$ is the set of correctly predicted examples of model $\mathcal{M}$. Figure \ref{fig:overlap} shows the overlap of models' predictions in the Situation dataset. We observe that Z-LaVI (w/o LM) has an obviously smaller overlap with the other models, while different language models have a big mutual overlap. This difference explains the substantial performance gain after exploiting visual imagination.
\medbreak
\noindent \textbf{\textsc{Recall} vs. \textsc{Synthesis}.} We ablate on the imagination methods and compare the performance of only using one of the methods. Table \ref{tab image method} demonstrates the performance on each dataset with different imagination methods. We can see that for the dataset with short inputs for imagination
(e.g., QA tasks), \textsc{Recall} is better than \textsc{Synthesis}. This is because 
short inputs of science QA datasets normally correspond to objects that exist in the real world and are easy to find on the web, such as \textit{mollusca} and \textit{porifera} shown in Figure \ref{fig:qa_examples} (a). However, for queries with long sentences (WSD and Topic Classification), the text inputs are too specific to match any real photo. Hence \textsc{Synthesis} is preferable.\footnote{We notice only 328 examples (out of 1789) of the Situation dataset have more than 10 images by \textsc{Recall}.} Figure \ref{fig:image_type} also indicates that the model prefers to choose \textsc{Recall} images for short input and tends to use \textsc{Synthesis} images when the input contains more tokens. We also find that without images, \model~has poor performance on all tasks, reflecting the necessity of imagination.

\begin{table}[!t]
\centering
\resizebox{7.7cm}{!}{
\begin{tabular}{ccccccc}
\Xhline{3\arrayrulewidth}
 & \textbf{QASC} & \textbf{SciQ} & \textbf{ARC} & \textbf{WSD} & \textbf{AG} & \textbf{Situ} \\ \hline
\# TOK   &  2.4    & 2.6    & 4.9 & 28.4  &  54.7  &  56.8    \\ \hline
\xmark  &  26.8    & 41.1    & 42.1 & 75.0  &  52.9  &  22.4    \\
\textsc{Recall}    &  \underline{32.6}    & \textbf{49.9}     &   \underline{48.5} & 80.3  & 52.5   &  21.1  \\
\textsc{Synthesis}  & 31.5 & 39.9 & 44.6 & \underline{81.5} & \textbf{71.6} & \textbf{34.2}\\
\textsc{Both}  & \textbf{32.7} & \underline{49.5} & \textbf{50.2} & \textbf{83.8} & \textbf{71.6} & \underline{33.4} \\\Xhline{3\arrayrulewidth}
 
\end{tabular}
}
\caption{The performance of Z-LaVI (w/o LM) with different imagination methods. \# TOK is the average number of tokens of text inputs in each dataset. \xmark~means no image is provided to the model, and we only use the text encoder of CLIP. \textsc{Recall} and \textsc{Synthesis} represent using image search and image generation, respectively. \textsc{Both} means combining the two methods.}
\label{tab image method}
\end{table}

\begin{figure}[!t]
\centering
    \includegraphics[width=7.7cm]{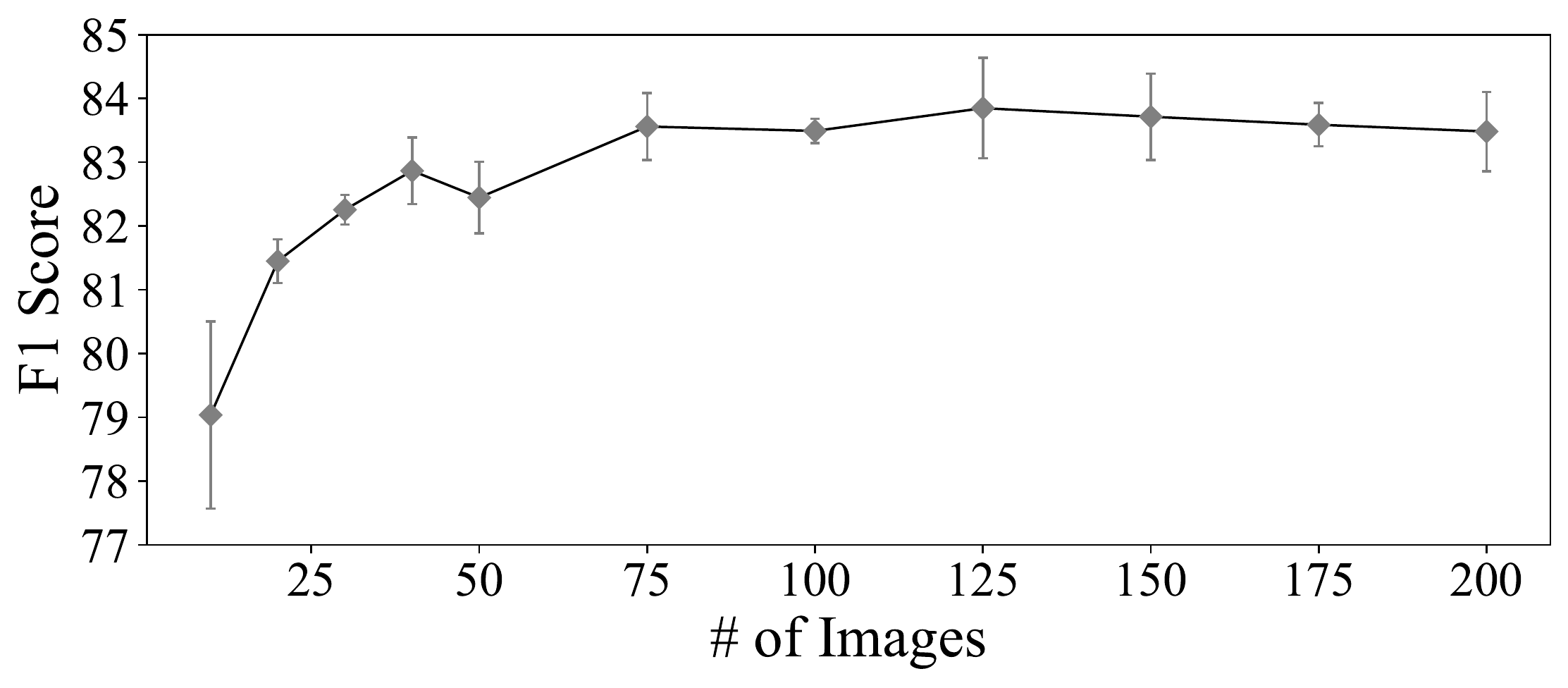}
    \caption{The performance on CoarseWSD-20 with the number of image candidates provided by imagination.}
    \label{fig:image_number}
\end{figure}

\begin{table*}[!t]
\centering
\resizebox{16cm}{!}{%
\begin{tabular}{lcccc|cccc|cccc}
\Xhline{3\arrayrulewidth}  
\multirow{3}{*}{\textbf{Model}} & \multicolumn{4}{c|}{\textbf{\textsc{Color}}} & \multicolumn{4}{c|}{\textbf{\textsc{Shape}}} & \multicolumn{4}{c}{\textbf{\textsc{Material}}}\\ \cline{2-13}
& \multicolumn{2}{c}{Original} & \multicolumn{2}{c|}{Z-LaVI} & \multicolumn{2}{c}{Original} & \multicolumn{2}{c|}{Z-LaVI} & \multicolumn{2}{c}{Original} & \multicolumn{2}{c}{Z-LaVI} \\
& $\rho$ & Acc@1 & $\rho$ & Acc@1 &  $\rho$ & Acc@1 &  $\rho$ & Acc@1 &  $\rho$ & Acc@1 &  $\rho$ & Acc@1 \\\hline
Random & -0.1 & 6.6 & - & - & 1.3 & 7.1 & - & - & -1.5 & 6.0 & - & -\\
BERT-L$^{\dagger}$  & 37.6 & 30.3 & - & - & 42.7 & 28.4 & - & - & 36.6 & 35.7 & - & - \\
Oscar-L$^{\dagger}$ & 31.8 & 17.1 & - & - & 40.0 & 38.1 & - & - & \textbf{39.2} & 40.5 & - & - \\
RoBERTa-L-mnli* & 35.3 & 25.1 & \underline{41.4} & \underline{38.2} & 41.7 & 66.4 & \underline{43.5} & \underline{67.9} & 28.3 & 34.9 & \underline{31.6} & \underline{37.7} \\
BART-L-mnli* & 34.6 & 27.5 & \underline{38.2} & \underline{32.6} & 41.5 & 68.6 & \underline{42.0} & \underline{69.3} & 30.5 & 37.0 & \underline{32.6} & \underline{40.5}\\
GPT-Neo-1.3B & 40.1 & 31.7 & \underline{47.4} & \underline{48.4} & 44.2 & 52.9 & \underline{46.1} & \underline{64.3} & 35.1 & 34.5 & \underline{36.4} & \underline{41.2} \\
GPT-Neo-2.7B & 41.3 & 29.3 & \underline{47.0} & \underline{43.6} & 43.5 & 50.0 & \underline{45.2} & \underline{62.1} & 35.5 & 30.6 & \underline{36.1} & \underline{37.7}\\
GPT-J-6B     & 44.3 & 38.0 & \underline{\textbf{49.6}} & \underline{50.9} & 46.9 & 65.7 & \underline{47.1} & \underline{70.0} & 38.5 & 42.3 & 37.9 & \underline{46.5}\\
OPT-30B & 41.9 & 41.3 & \underline{48.1} & \underline{\textbf{52.1}} & 46.4 & 60.0 & \underline{\textbf{48.4}} & \underline{\textbf{72.1}} & 38.3 & 44.7 & 38.1 & \underline{\textbf{47.5}}\\
SimCSE & 30.7 & 34.8 & \underline{36.3} & \underline{40.4} & 30.1 & 28.6 & \underline{34.7} & \underline{40.7} & 24.6 & 26.4 & \underline{29.2} & \underline{33.5}\\
SBERT* & 27.6 & 26.5 & \underline{38.2} & \underline{40.6} & 20.3 & 13.6 & \underline{33.6} & \underline{35.0} & 24.1 & 22.5 & \underline{30.4} & \underline{34.9}\\ \hline
Z-LaVI w/o LM & - & - & 37.2 & 39.4 & - & - & 33.8 & 32.1 & - & - & 24.9 & 32.7 \\ \Xhline{3\arrayrulewidth}
\end{tabular}
}
\caption{Zero-shot probing on the three relation types (\textsc{Color}, \textsc{Shape} and \textsc{Material}) in ViComTe \cite{zhang-etal-2022-visual} dataset. We report the average Spearman correlation ($\rho$) and top-1 accuracy (Acc@1).}
\label{tab: vicomte}
\end{table*}

\paragraph{Performance vs. Image Quantities.} We combine \textsc{Recall} and \textsc{Synthesis} to imagine 200 image candidates. We wonder whether the number of imaginations impacts the \model's performance. Figure \ref{fig:image_number}
 reports \model's performance on CoarseWSD-20 versus the number of images. We observe that \model’s F1 score increases with a higher number of images. While the improvement is marginal when the number is higher than 125.

\paragraph{\model~supplements visual commonsense knowledge.} To further validate \model~helps to mitigate reporting bias problems of language models, we conduct experiments on ViComTe \cite{zhang-etal-2022-visual}, which is a commonsense knowledge dataset containing different types of properties for over 5000 subjects, e.g., the subject ``\textit{egg}'' has the property (object) ``\textit{oval}''. We investigate three relation types (\textsc{Color}, \textsc{Shape}, and \textsc{Material}) and report the results on the test set (see Table \ref{tab: vicomte stat} for details). We select the BERT-large, and Oscar-large \cite{li2020oscar} as the baselines of which the results are directly obtained from \citet{zhang-etal-2022-visual}.\footnote{We also include the random baseline by assigning a number between 0 and 1 for each class by chance. We iterate the random runs 7 times and report the average performance.} For a fair comparison, we adopt the same set of seven prompt templates provided by \citet{zhang-etal-2022-visual} and report the average performance over these prompts. Table \ref{tab: vicomte} demonstrates the performance of \model~with language models. We can see \model~continue to consistently boost the performance of language models and outperform the baselines with significant margins. The results on ViComTe indicate \model~is a promising way to overcome the reporting bias of language models on visual commonsense. 

\begin{table}[!t]
\centering
\centering
\resizebox{7.7cm}{!}{%
\begin{tabular}{lccc}
\Xhline{3\arrayrulewidth}  
\textbf{Relation}      &    \textbf{\# Objs} & \textbf{\# Test Subjs} & \textbf{Ex (subj, obj) pair}  \\ \hline
\textsc{Color}         &    12 & 574 &  (leave, green)                        \\
\textsc{Shape}         &    12 & 140 &  (egg, oval)                           \\
\textsc{Material}      &    18 & 284 &  (mug, glass)                          \\ \Xhline{3\arrayrulewidth}  
\end{tabular}
}
\caption{Statistics of ViComTe dataset.}
\label{tab: vicomte stat}
\vspace{-.2cm}
\end{table}

\begin{figure}[!t]
\centering
    \includegraphics[width=7.7cm]{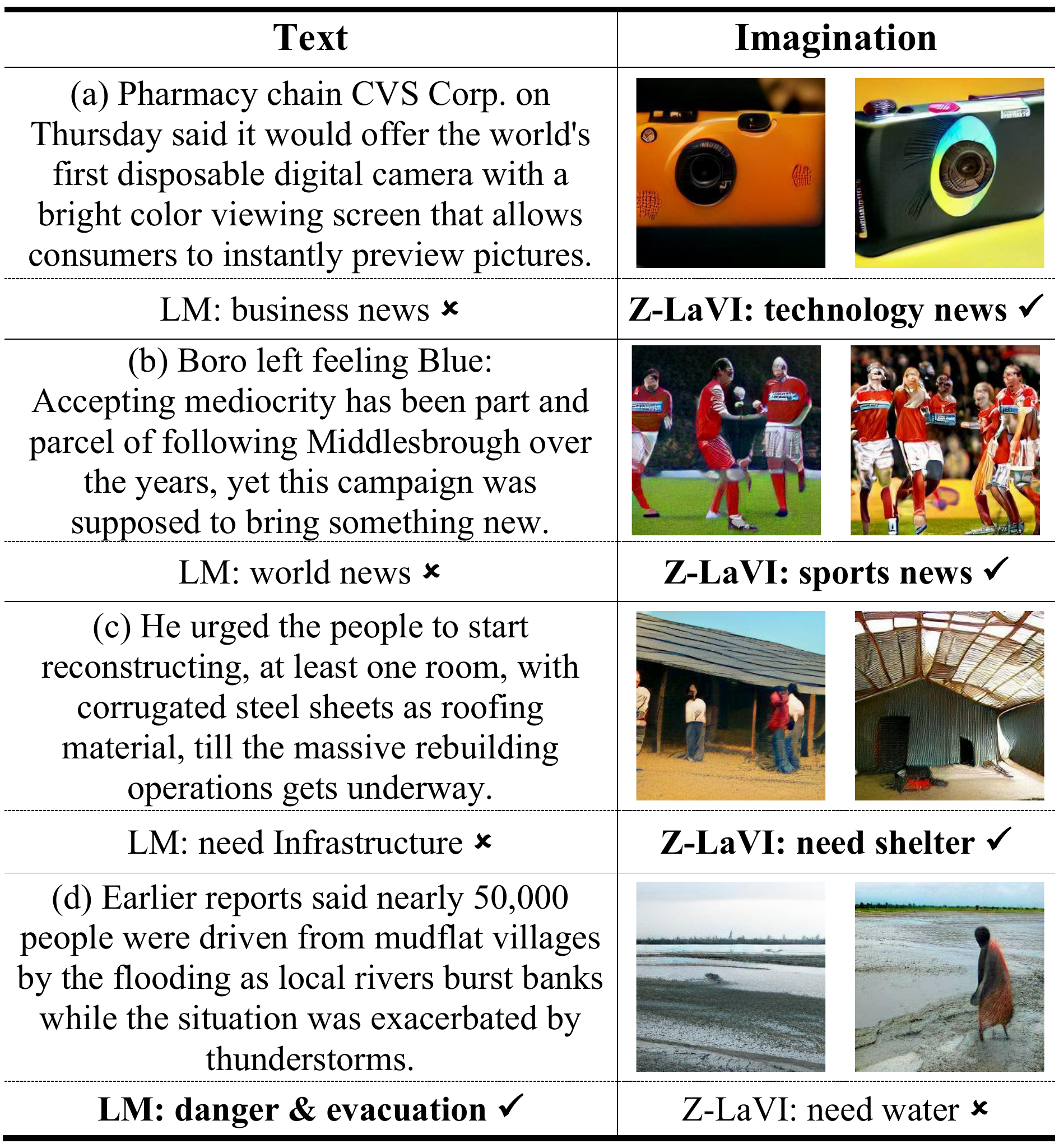}
    \caption{Qualitative examples from AG-News (a, b) and Situation (c, d).}
    \label{fig:tc examples}
    \vspace{-.15cm}
\end{figure}

\paragraph{Qualitative Examples.} Figure \ref{fig:tc examples} shows some qualitative examples from the two topic classification datasets. We observe \model~can effectively correct language models' prediction with more straightforward visual signals. However, we also notice that \model~fails on examples that cannot be solved by correlation, e.g., \model~wrongly relates \textit{flooding} with the situation of \textit{need water}. More examples are provided in the Appendix \ref{appendix: qual examples}.

\section{Related Work}
\textbf{Visually Grounded Representation Learning} Several studies have focused on learning visually grounded word or sentence representations. To learn better word embeddings, \citet{kiros-etal-2018-illustrative} introduce Picturebook that encodes images as vectors by querying all vocabulary words through Google image search. 
\citet{lazaridou2015combining} optimize a multimodal skip-gram model, where visual information is presented together with the corpus contexts to produce word embeddings.
\citet{zablocki2018learning} leverage the visual context of objects to learn multimodal word embeddings. 
With respect to visually grounded sentence embeddings, previous work develops several strategies to enrich sentence representations with visual information, such as using the given sentence as captions to get image features \citep{kiela-etal-2018-learning}, capturing both cluster information and perceptual information in grounded space \citep{bordes-etal-2019-incorporating}, or exploiting multimodal contrastive learning objective \cite{zhang2022mcse}.
\citet{zhang2021semi} retrieves images from COCO \cite{lin2014microsoft} as augmented visual features for language models.
\citet{lu2022imagination} augment sentence embeddings with VQGAN-\citep{esser2021taming} generated images and fine-tune them on GLUE Benchmark \citep{wang2018glue}. \citet{liu-etal-2022-things} probes the spatial commonsense knowledge (sizes, positions) of language models and vision-language models through image generation.
\medbreak
\noindent \textbf{Vision-Language Pretraining Models} 
To connect vision and language semantics, a line of work on multimodal masked language models \cite{li2019visualbert, tan2019lxmert, NEURIPS2019_c74d97b0, Su2020VL-BERT} explores vision-language pretraining and achieves SOTA fine-tuning performance on multimodal benchmarks. \citet{tsimpoukelli2021multimodal} freeze a language model parameters to generate the appropriate caption by encoding each image into the embedding space to inject visual knowledge into PLMs. To retrain knowledge in both vision and language pretrained models, Flamingo \citep{alayrac2022flamingo} freezes both pretrained models and brings in additional model components to do visually-conditioned autoregressive text generation. \citet{tan-bansal-2020-vokenization} retrieve related images as vokens (visualized tokens) and then process large language corpora (e.g., Wikipedia) into voken-prediction tasks. FLAVA \citep{singh2022flava} is an alignment model that pretrains on both unimodal and multimodal data while optimizing cross-modal ``alignment'' objectives and multimodal fusion objectives. Unified-IO \cite{lu2022unified} is a general-purpose model which can perform a wide range of vision, language, and multimodel tasks by unifying the inputs and outputs as sequences.

\vspace{-.1cm}
\section{Conclusion}
\vspace{-.1cm}
In this paper, we propose a novel approach, Z-LaVI, to alleviate the reporting bias problem of pretrained language models and enhance their zero-shot inference ability. We develop two complementary visual imagination mechanisms, i.e., \textsc{Recall} that aims to retrieve existing objects or scenes and \textsc{Synthesis} that generates nonexistent ones. Experiments on a wide range of language tasks show that our approach can significantly outperform existing zero-shot language models, pointing towards a promising direction to solve an unseen language task with visual imagination.

\section{Limitations}
Our experiments apply DALL$\cdot$E-mini for synthesizing the images, but the quality and resolution of the generated images are still low, which can be the factor limiting \model's performance. However, the recent breakthroughs of DALLE$\cdot$E-2 \cite{ramesh2022hierarchical}, Imagen, \cite{saharia2022photorealistic} and the open-sourced Stable Diffusion \cite{rombach2022high} give us hope to obtain more realistic images and thus further unleash the potential of \model. The negative results on ARC-C reveal the lack of complex reasoning ability in the current zero-shot vision-text model. At the same time, the success of Flamingo \cite{alayrac2022flamingo} on few-shot multi-modal tasks lets us sense the possibility of applying the framework of \model~with these powerful visual language models to solve broader language tasks. We can foresee the bright future of our method once these powerful resources are publicly available. In this paper, we focus on the zero-shot settings, and thus it is difficult to design a more effective approach to ensemble the language and vision without training data. However, when few-shot examples are available, it is possible to learn a mechanism to automatically calibrate the weights of imagination depending on the input examples.

In addition, the image generation model is trained on unfiltered data on the web, which may leak personal information such as the human face, etc. The generation model may also be biased toward stereotypes against minority groups. Furthermore, compared to language models, our approach requires extra resources such as an image search engine, pretrained text-to-image generation model, etc., which will increase the implementation cost. Finally, we evaluated our method in English datasets only, and we plan to incorporate other languages in the future with the help of multilingual multi-modal models \cite{huang-etal-2021-multilingual}. 

\bibliography{anthology,custom}
\bibliographystyle{acl_natbib}
\clearpage

\appendix
\section{Implementation Details}
\subsection{Prompt Selection}
We use the intuitive prompt templates of the three main tasks for each model shown in Table \ref{tab: prompt}, and we do not tune the prompt for each dataset. For ViComTe dataset, we reuse the original 7 prompts provided by \citet{zhang-etal-2022-visual}. 

\subsection{Model Parameters}
We include the number of parameters of all models we use in Table \ref{tab n parameters}. We also list the ensemble weight $w$ based on the relative sizes between the two models:
\begin{equation}
    w = \text{sigmoid}\left(\frac{\mathcal{P}_{\text{VI}}}{\mathcal{P}_{\text{La}}}\right) = \frac{1}{1 + e^{-\mathcal{P}_{\text{VI}} / \mathcal{P}_{\text{La}}}}
\end{equation}


\begin{table}[!htbp]
\centering
\small
\begin{tabular}{ccc}
\Xhline{3\arrayrulewidth}      
\textbf{Model}        & \textbf{\# of Parameters} & \textbf{Ensemble weight} \\ \hline
CLIP         & 150 M            & -               \\
BART-L       & 400 M            & 0.59            \\
RoBERTa-L    & 355 M            & 0.60            \\
SBERT        & 110 M            & 0.80            \\
SimCSE       & 355 M            & 0.60            \\
GPT-Neo-1.3B & 1.3 B            & 0.53            \\
GPT-Neo-2.7B & 2.7 B            & 0.51            \\
GPT-J-6B     & 6 B              & 0.51            \\
OPT-30B      & 30 B             & 0.50            
\\\Xhline{3\arrayrulewidth}      
\end{tabular}
\caption{The number of parameters of each model and the ensemble weights calculated by equation \ref{eq: ensemble weight}.}
\label{tab n parameters}
\end{table}

\subsection{Other Details}
\textbf{Predicted scores of prompt-based approach.} The huggingface API can calculate the loss of the provided input, and we take the reciprocal of the loss as the prediction score in practice.
\medbreak
\noindent \textbf{Speed of Bing Image Search.} The efficiency of downloading images from Bing depends on your internet speed and the API plan you select. In our case, downloading 200 images of a query takes less than 10 seconds. The script of Bing Image Search is provided in our GitHub repo.
\medbreak
\noindent \textbf{Requirements for computing resources.} Our approach is designed for zero-shot scenarios and does not involve training, thus most of the experiments can be conducted on a single GPU with more than 20GB of memory. The biggest challenge is to deploy the OPT-30B, which requires 70GB of memory. We successfully deploy OPT-30B using 4 P40 GPUs and released the code to implement OPT-30B across multiple GPUs in our GitHub repo.

\section{Additional Analysis}
\subsection{Relative Improvement of Each Word in Word Sense Disambiguation}
To quantify the ability of \model~to disambiguate different words, we compute the relative improvement (\texttt{RI}) of each word by comparing the F1 score of SBERT with that of \model:

\begin{equation}
    \texttt{RI} = \frac{\text{F1}_{\text{\model}} - \text{F1}_{\text{SBERT}}}{\text{F1}_{\text{SBERT}}}
\end{equation}

\begin{figure*}[!t]
\centering
    \includegraphics[width=15.5cm]{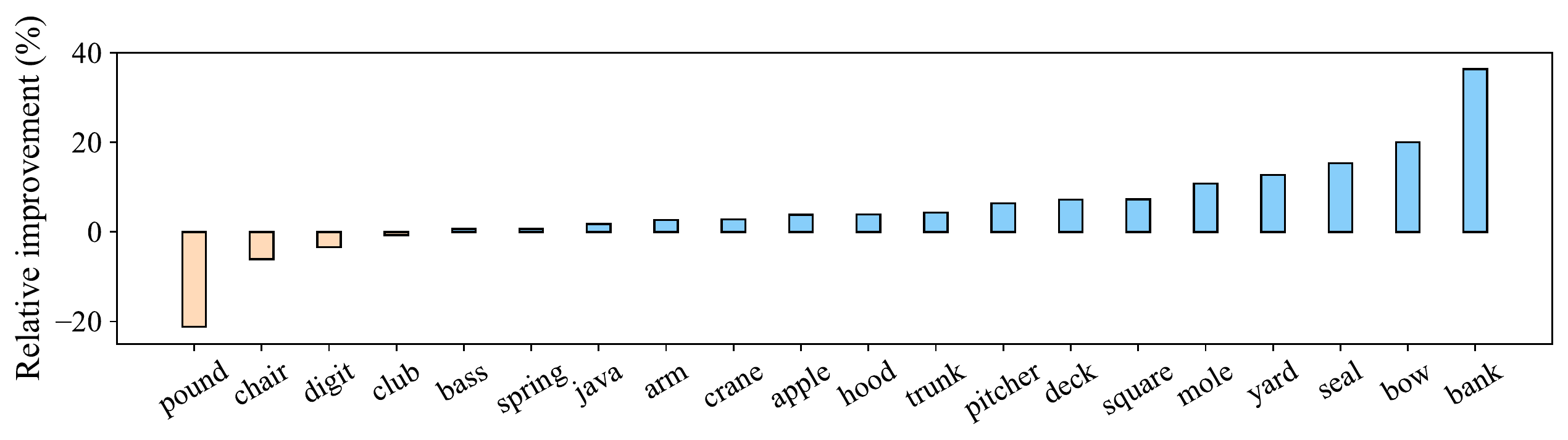}
    \caption{Relative improvement (F1) of the 20 words in CoarseWSD20.}
    \label{fig:wsd word}
\end{figure*}

As shown in Figure \ref{fig:wsd word}, \model~improves the F1 score of 16 words (out of 20), while for the rest of four words (\textit{pound}, \textit{chair}, \textit{digit}, \textit{club}), \model~hurts the performance. Furthermore, we notice those four words all contain abstract senses (see Figure \ref{fig:wsd examples 1}) which are difficult to imagine, e.g., \textit{pound} has two senses-\textit{currency} and \textit{mass}, which are both units and difficult to illustrate in images. Based on this finding, we think future work can design a dynamic ensemble method to calibrate the weights conditioned on the input.

\subsection{Overlap of Incorrectly Predicted Examples}
\begin{figure}[!htbp]
\centering
    \includegraphics[width=7cm]{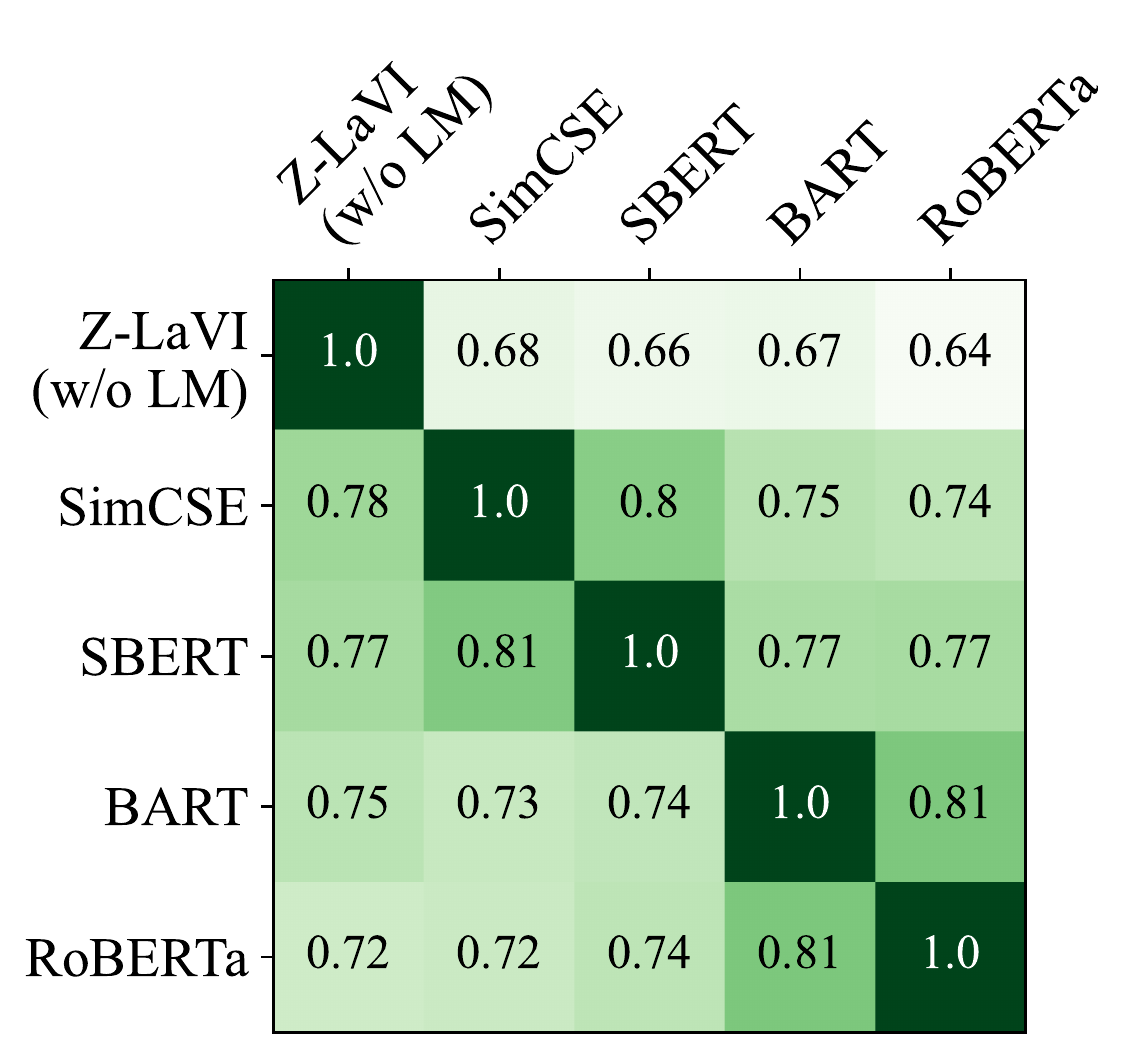}
    \caption{The overlap of incorrectly predicted examples between each pair of models in the Situation dataset.}
    \label{fig:overlap_incorrect}
\end{figure}

In Figure \ref{fig:overlap}, we show that \model~w/o LM has a smaller overlap with other language models on \textbf{correctly} predicted examples. To tell a complete story of the different output distribution between \model~and language models, we compute the overlap of \textbf{incorrectly} predicted examples shown in Figure \ref{fig:overlap_incorrect}. We can see that \model~has a smaller overlap on incorrectly predicted ones.

\begin{table*}[ht!]
\centering
\small
\begin{tabular}{cccc}
\Xhline{3\arrayrulewidth}  
\textbf{Model} & \textbf{Word Sense Disambiguation} & \textbf{Question Answering} & \textbf{Topic Classification} \\ \hline
\begin{tabular}[c]{@{}c@{}}GPT\\ NLI\end{tabular} & \begin{tabular}[c]{@{}c@{}}{[}SENTENCE{]} The {[}TARGET WORD{]} \\ mentioned before means {[}SENSE NAME{]}.\end{tabular} & \begin{tabular}[c]{@{}c@{}}Question: {[}QUESTION{]} \\ The answer is {[}ANSWER{]}.\end{tabular} & \begin{tabular}[c]{@{}c@{}}{[}SENTENCE{]} This example \\ is {[}CLASS NAME{]}.\end{tabular} \\ \hline
\begin{tabular}[c]{@{}c@{}}SimCSE\\ SBERT\end{tabular}   & (SENTENCE, DEFINITION)   & (QUESTION, ANSWER) & \begin{tabular}[c]{@{}c@{}}(SENTENCE, This example \\ is {[}CLASS NAME{]}.)\end{tabular}    \\ \hline
CLIP    & (SENTENCE, DEFINITION)     & (QUESTION, ANSWER) & \begin{tabular}[c]{@{}c@{}}(SENTENCE, A news image \\ of {[}CLASS NAME{]}.)\end{tabular}   \\\Xhline{3\arrayrulewidth}  
\end{tabular}
\caption{The prompts we use for each model on WSD, QA, and Topic classification.}
\label{tab: prompt}
\end{table*}
\begin{table*}[!htbp]
\centering
\small
\begin{tabular}{ccc}
\Xhline{3\arrayrulewidth}
\textbf{\textsc{Color}} & \textbf{\textsc{Material}} & \textbf{\textsc{Shape}}  \\ \hline
\begin{tabular}[c]{@{}c@{}}{[}SUBJ{]} can be of color {[}OBJ{]}.\\ {[}SUBJ{]} has color {[}OBJ{]}.\\ The color of {[}SUBJ{]} can be {[}OBJ{]}.\\ The color of the {[}SUBJ{]} is {[}OBJ{]}.\\ {[}SUBJ{]} is {[}OBJ{]}.\\ This {[}SUBJ{]} is {[}OBJ{]}.\\ {[}SUBJ{]} is of color {[}OBJ{]}.\end{tabular} & \begin{tabular}[c]{@{}c@{}}{[}SUBJ{]} is made of {[}OBJ{]}.\\ {[}SUBJ{]} can be made of {[}OBJ{]}.\\ {[}SUBJ{]} is made from {[}OBJ{]}.\\ {[}SUBJ{]} can be made from {[}OBJ{]}. \\ {[}SUBJ{]} is {[}OBJ{]}.\\ This {[}SUBJ{]} is {[}OBJ{]}.\\ {[}SUBJ{]} is made by using {[}OBJ{]}.\end{tabular} & \begin{tabular}[c]{@{}c@{}}{[}SUBJ{]} can be shape of {[}OBJ{]}.\\ {[}SUBJ{]} has shape of {[}OBJ{]}.\\ {[}SUBJ{]} is of shape {[}OBJ{]}.\\ The shape of {[}SUBJ{]} can be {[}OBJ{]}.\\ The shape of the {[}SUBJ{]} is {[}OBJ{]}.\\ {[}SUBJ{]} is {[}OBJ{]}.\\ This {[}SUBJ{]} is {[}OBJ{]}.\end{tabular} \\
\Xhline{3\arrayrulewidth}
\end{tabular}
\caption{The prompts we use for ViComTe which are provided by \citet{zhang-etal-2022-visual}.}
\end{table*}

\subsection{\textsc{Recall} vs. \textsc{Synthesis} on ViComTe}
We also ablate on the imagination methods on ViComTe shown in Table \ref{tab: vicomte ablate}. As mentioned before, \textsc{Recall} is preferred when the text input is short. This finding still holds for ViComTe where the text inputs are single words. We notice \textsc{Recall} performs better than \textsc{Synthesis} except for the \textsc{Color} relation. We find the images from \textsc{Synthesis} are more prototypical in colors than the ones from \textsc{Recall}. (See examples in Figure \ref{fig:vicomte examples})
\begin{table}[!htbp]
\centering
\resizebox{7.5cm}{!}{%
\begin{tabular}{lcccc}
\Xhline{3\arrayrulewidth}  
  $\rho$/Acc@1       & \textbf{\textsc{Recall}} & \textbf{\textsc{Synthesis}} & \textbf{\textsc{Both}} \\ \hline
\textsc{color}       &  35.5/35.7     &   \underline{36.4}/\underline{38.5}    &     \textbf{37.2}/\textbf{39.4}             \\
\textsc{shape}       &  \textbf{37.1}/\textbf{49.3}     &   27.9/31.3    &     \underline{33.9}/\underline{32.1}             \\
\textsc{material}    & \textbf{29.9}/\textbf{34.9}      &   23.4/29.6    &  \underline{24.9}/\underline{32.7}     \\\Xhline{3\arrayrulewidth}  
\end{tabular}
}
\caption{The average Spearman correlation (left) and top-1 accuracy (right) of \model~w/o LM with different imagination methods. The highest number of each row is \textbf{bolded} and the second-best one is \underline{underlined}.}
\label{tab: vicomte ablate}
\end{table}

\section{Qualitative Examples} \label{appendix: qual examples}
\begin{figure*}[!t]
\centering
    \includegraphics[width=15.5cm]{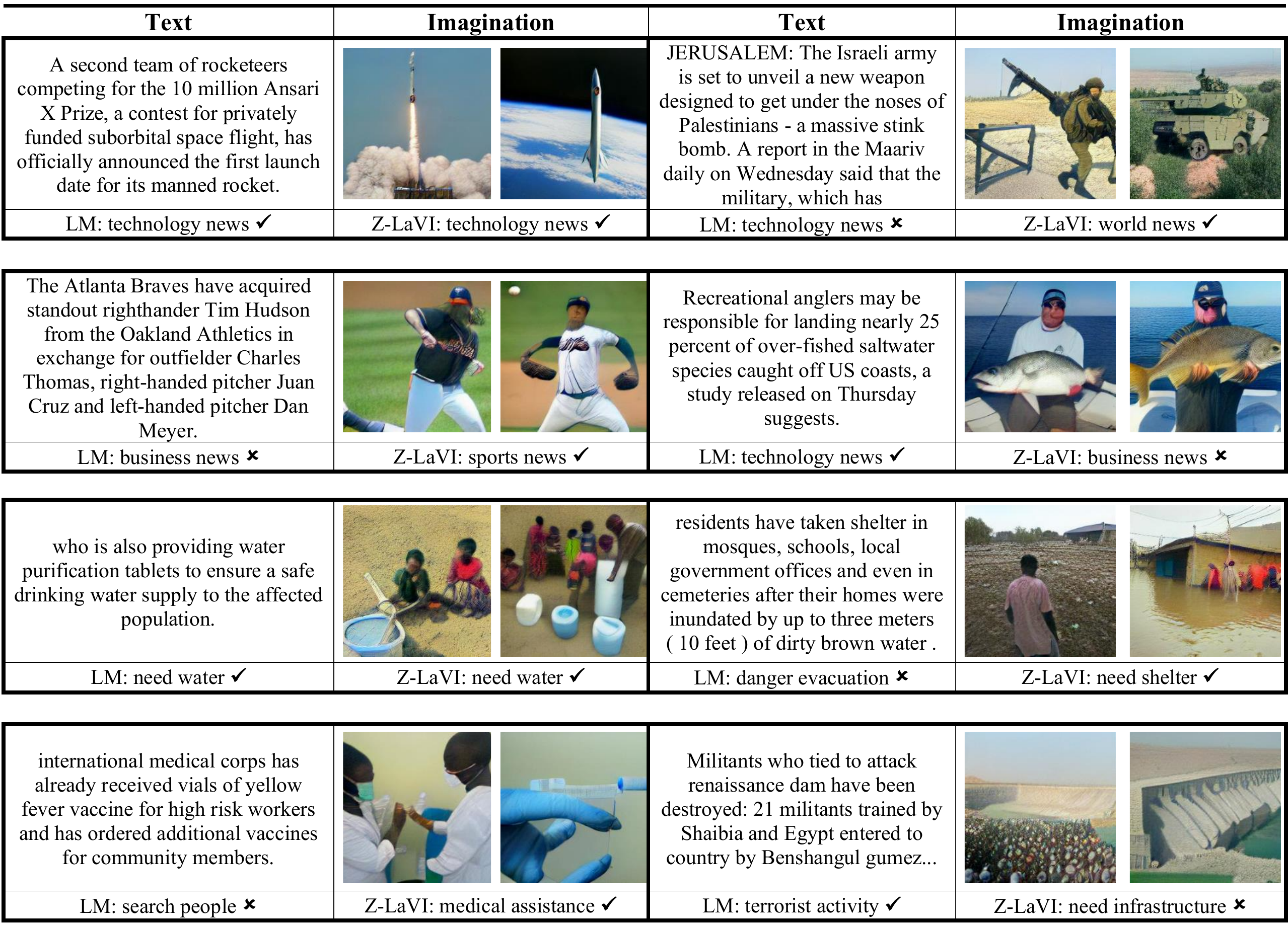}
    \caption{Qualitative examples from AG-News (top-4) and Situation (bottom-4).}
    \label{fig:tc examples full}
\end{figure*}

\begin{figure*}[!t]
\centering
    \includegraphics[width=15.5cm]{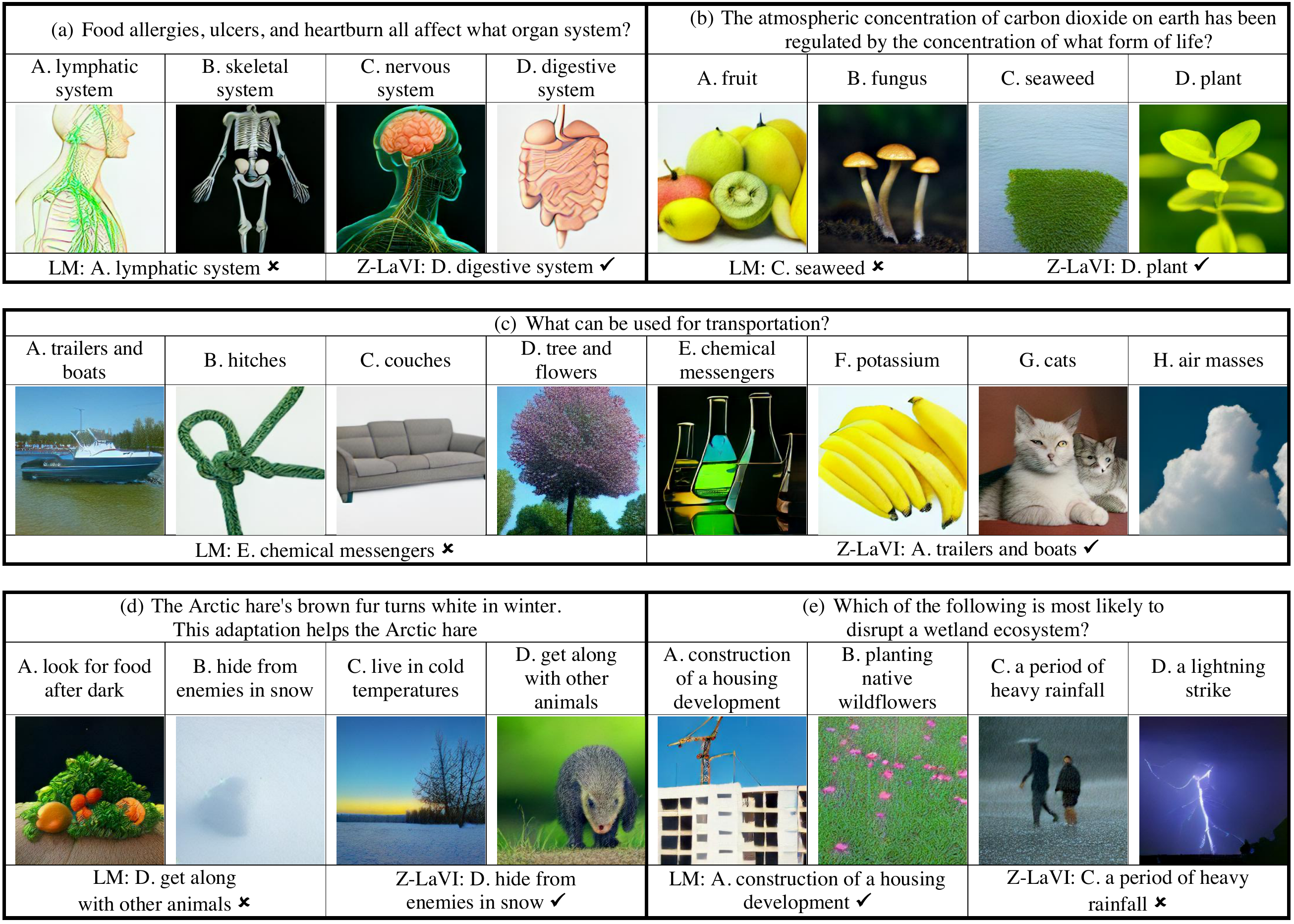}
    \caption{Qualitative examples from SciQ (a, b), QASC (c), ARC-E (d), and ARC-C (e). (e) is a typical failure case of CLIP because multimodal models are not good at understanding negation \cite{jimenez-etal-2022-carets}.}
    \label{fig:qa examples full}
\end{figure*}

\begin{figure*}[!t]
\centering
    \includegraphics[width=15.5cm]{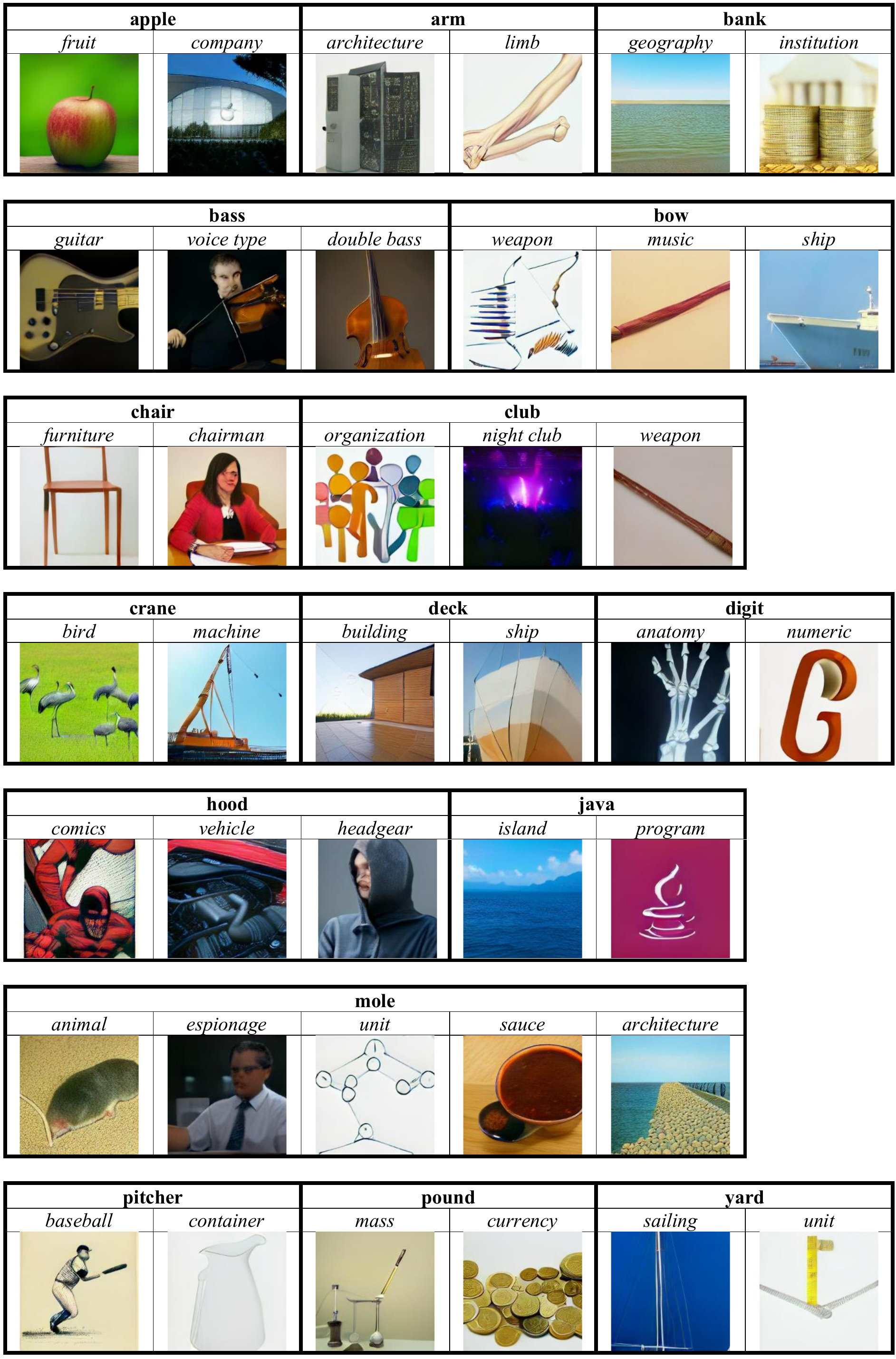}
    \caption{Qualitative examples from CoarseWSD20. We randomly sample 1 image (\textsc{Synthesis}) for each sense.}
    \label{fig:wsd examples 1}
\end{figure*}

\begin{figure*}[!t]
\centering
    \includegraphics[width=15.5cm]{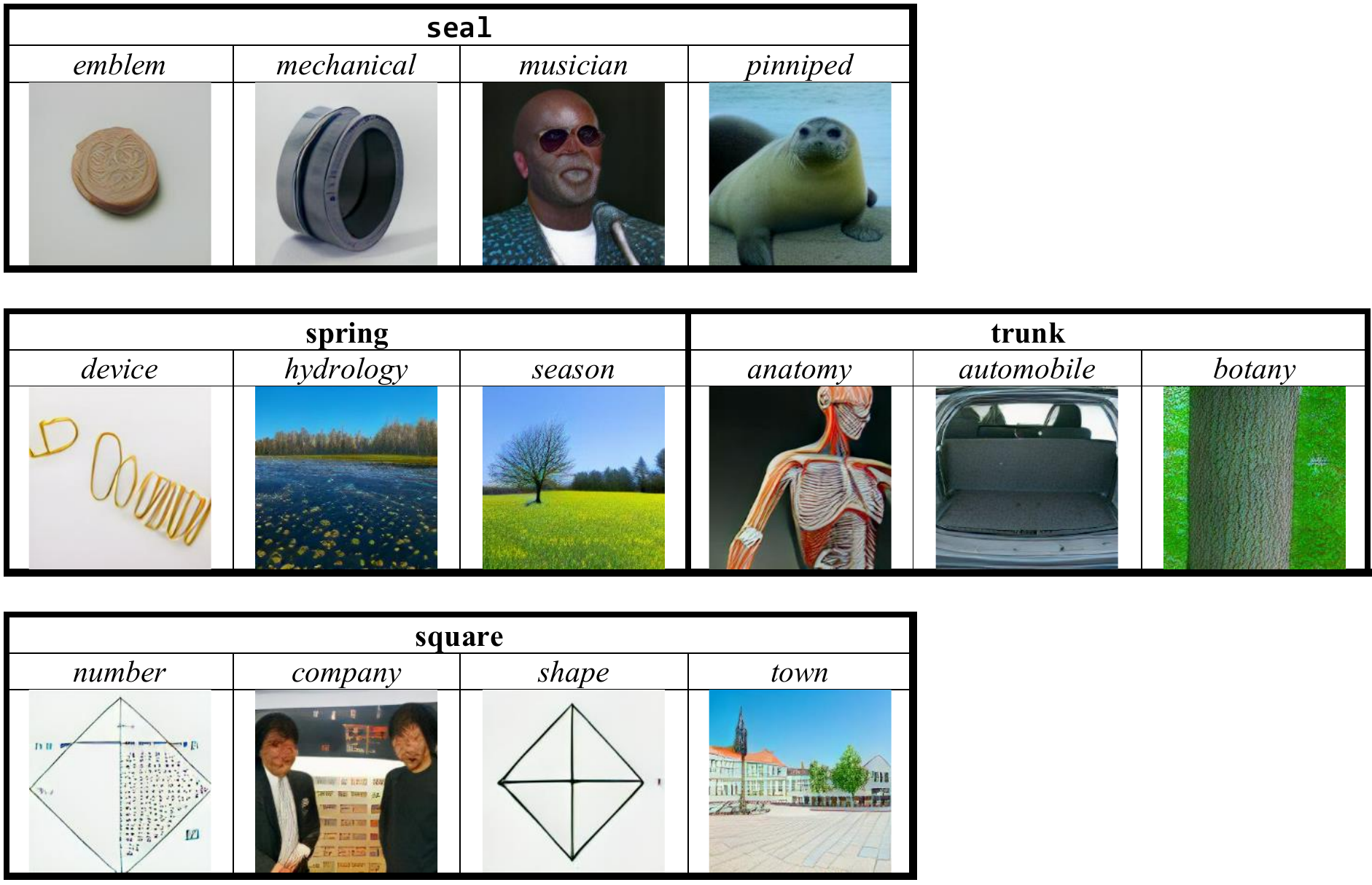}
    \caption{Qualitative examples from CoarseWSD20. (Continued)}
    \label{fig:wsd examples 2}
\end{figure*}

\begin{figure*}[!t]
\centering
    \includegraphics[width=15.5cm]{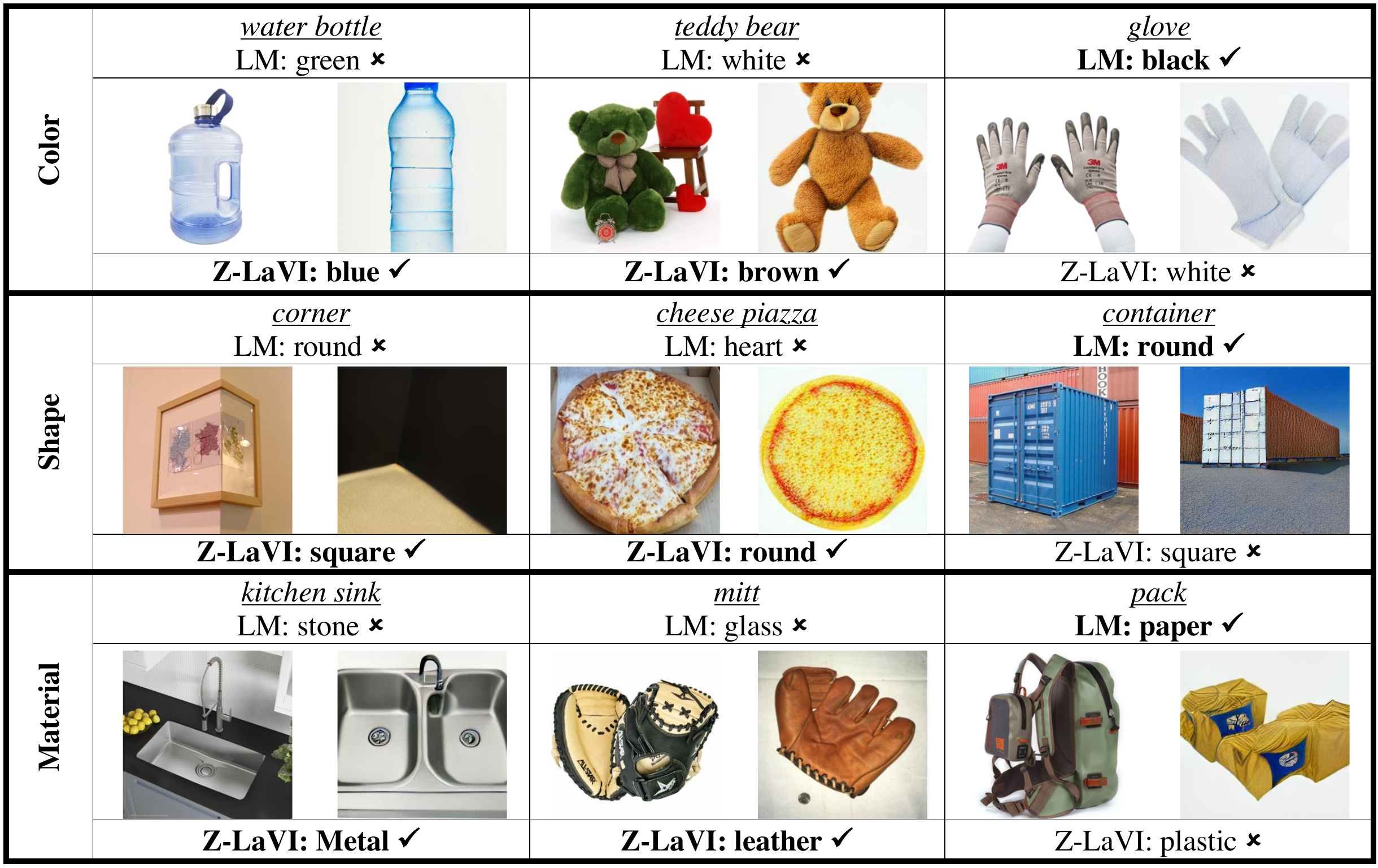}
    \caption{Qualitative examples from ViComTe. We randomly sample three subjects for each relation type (\textsc{Color}, \textsc{Shape}, and \textsc{Material}). For each subject, we present two images, one from \textsc{Recall} (left) and another one from \textsc{Synthesis} (right).}
    \label{fig:vicomte examples}
\end{figure*}

Figure \ref{fig:tc examples full} demonstrates the qualitative examples from the two topic classification dataset. Figure \ref{fig:qa examples full} shows the QA examples. Figure \ref{fig:wsd examples 1}, \ref{fig:wsd examples 2} show the \textsc{Synthesis} image of each word sense from CoarseWSD20. Figure \ref{fig:vicomte examples} demonstrates the images (\textsc{Synthesis} and \textsc{Recall})of different subjects from ViComTe. In addition, we release all the \textsc{Synthesis} images generated by DALL$\cdot$E-mini which can be downloaded from our GitHub repo.

\end{document}